\documentclass[final,3p,times,twocolumn]{elsarticle}
\usepackage[utf8]{inputenc}

\usepackage{natbib}

\usepackage{microtype}
\usepackage{graphicx}
\usepackage{subfigure}
\usepackage{booktabs} 

\usepackage{subfigure}
\usepackage{graphicx}
\usepackage{amsmath}
\usepackage{amsfonts}
\usepackage{xcolor}
\usepackage{caption}
\usepackage{mathtools}
\usepackage{bm}
\usepackage{soul}
\usepackage{booktabs}
\usepackage{hyperref}
\usepackage{wrapfig}
\usepackage{listings}
\usepackage{setspace}
\usepackage{multirow}
\usepackage{multicol}

\usepackage{tcolorbox}

\usepackage{algorithm}
\usepackage{algpseudocode}

\usepackage{titlesec}

\usepackage{makecell, multirow, threeparttable}

\newcommand{\para}[1]{\vspace{.05in}\noindent\textbf{#1}}

\journal{Neural Networks}

\title{Democratizing Large Language Model-Based Graph Data Augmentation \\ via Latent Knowledge Graphs}

\begin{document}

\author[inst1]{Yushi Feng \corref{cor1}}

\affiliation[inst1]{organization={Department of Statistics and Actuarial Science, School of Computing and Data Science, The University of Hong Kong},
            city={Hong Kong SAR},
            country={China}
            }

\author[inst1]{Tsai Hor Chan \corref{cor1}}
\author[inst1]{Guosheng Yin}
\author[inst1]{Lequan Yu \corref{cor2}}
\cortext[cor1]{Equal contributions}
\cortext[cor2]{Corresponding author. Email address: lqyu@hku.hk}

\begin{abstract}
  Data augmentation is necessary for graph representation learning due to the scarcity and noise present in graph data.
  Most of the existing augmentation methods overlook the context information inherited from the dataset as they rely solely on the graph structure for augmentation.
  Despite the success of some large language model-based (LLM) graph learning methods, they are mostly white-box which require access to the weights or latent features from the open-access LLMs, making them difficult to be democratized for everyone as existing LLMs are mostly closed-source for commercial considerations.
  %
  To overcome these limitations, we propose a black-box context-driven graph data augmentation approach, with the guidance of LLMs --- \textbf{DemoGraph}.
    Leveraging the text prompt as context-related information, we task the LLM with generating knowledge graphs (KGs), which allow us to capture the structural interactions from the text outputs.
    We then design a dynamic merging schema to stochastically integrate the LLM-generated KGs into the original graph during training.
    To control the sparsity of the augmented graph, we further devise a granularity-aware prompting strategy and an instruction fine-tuning module, which seamlessly generates text prompts according to different granularity levels of the dataset.
    Extensive experiments on various graph learning tasks validate the effectiveness of our method over existing graph data augmentation methods. 
    Notably, our approach excels in scenarios involving electronic health records (EHRs), which validates its maximal utilization of contextual knowledge, leading to enhanced predictive performance and interpretability.
\end{abstract}

\maketitle

\section{Introduction}

Graph representation learning has received increasing attention in recent years. 
It achieves great success in solving tasks where relational features are important, such as recommendation systems \citep{cai2023lightgcl, shi2018HeRec}, citation networks \citep{hu2020OGB}, and medical records analysis \citep{choi2018mime, ma2018kame}. 
However, the scarcity and noise present in graph data pose great challenges for effective graph learning, necessitating the development of graph data augmentation algorithms. 

Existing graph data augmentation methods focus on graph structures for data augmentation,  such as randomly dropping nodes or edges, adding Gaussian noise to the node or edge attributes, or applying graph-based transformations such as sub-sampling and node permutation. 
While these methods have demonstrated some successes in graph representation learning scenarios, they do not consider the \textit{context} or \textit{attributes} associated with the graph data.
This prompts some recent works \cite{jiang2023graphcare, tang2023graphgpt, he2023harnessing, huang2024higher, wei2024llmrec, west2021symbolic, zhang2022greaselm} which leverage LLM for graph representation learning.
\begin{figure}
    \centering
    \includegraphics[width=0.4\textwidth]{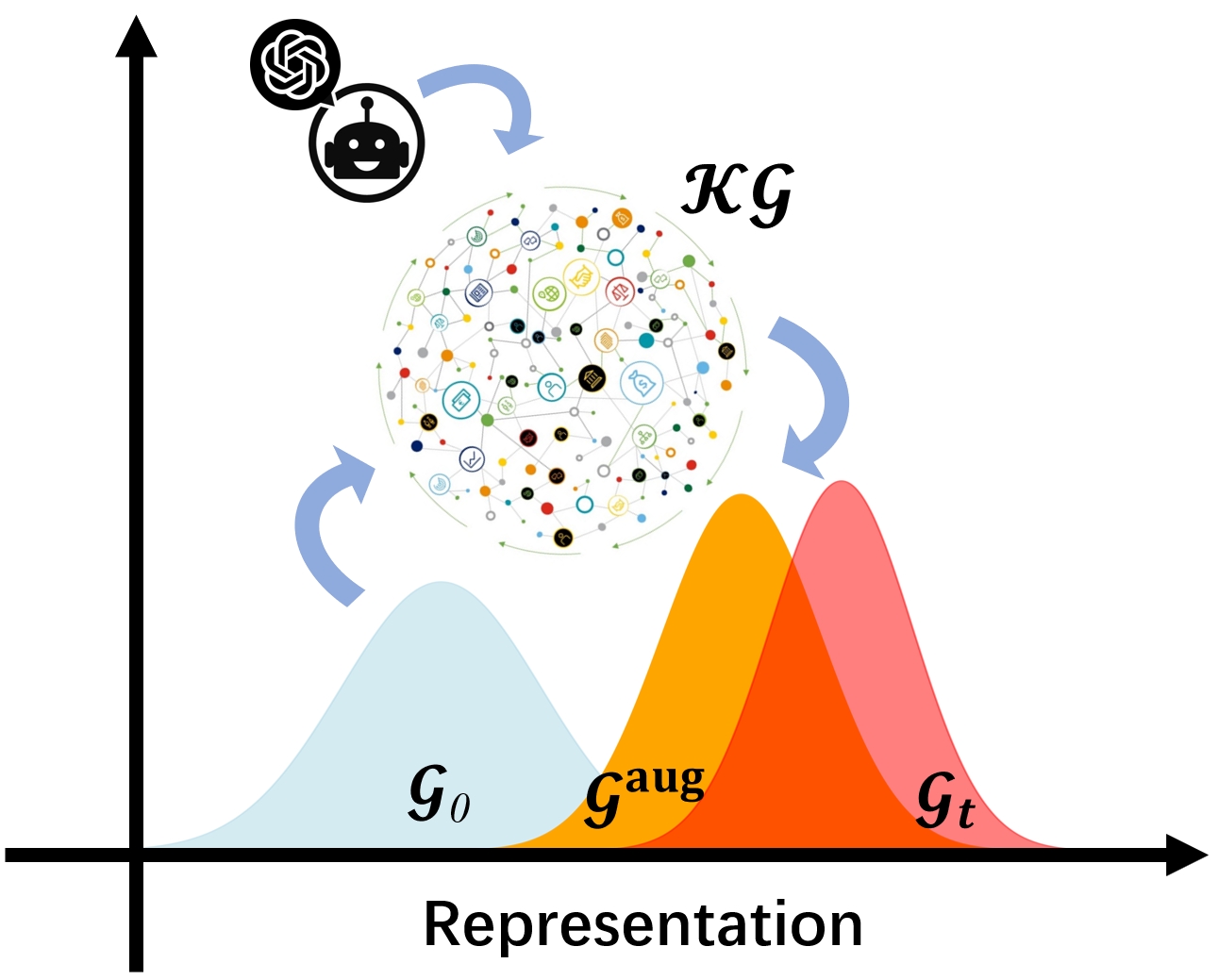}
    \caption{Schematic illustration of the feature distribution of original graph $\mathcal{G}_0$ from observations and $\mathcal{G}^{\text{aug}}$, which represents the augmented graph for $\mathcal{G}_0$ after merging the context knowledge in terms of $\mathcal{KG}$. After performing graph data augmentation with LLM-guided DemoGraph, $\mathcal{G}^{\text{aug}}$ is closer to the true representation $\mathcal{G}_t$.}
    \label{fig:ehr_hg_example}
\end{figure}
Despite their success, they are mostly white-box which require access to the weights or latent features from the LLMs, making them difficult to be democratized as existing LLMs are mostly closed-source for commercial considerations.
As a result, the resulting augmented graph becomes less identifiable due to a lack of contextual guidance.
Furthermore, most of these augmentation methods leverage in-domain knowledge under a close-world setting, which does not borrow the vast repositories of knowledge in the open world.
Additionally, the sparsity of the augmented graph is not well studied, although some methods, such as DropEdge, attempt to sparsify the graph for augmentation.
Without proper sparsity control, the augmented graph would be over-sparsified and likely reduced to trivial graphs (i.e., uninformative graphs).
These limitations pop the necessity of developing a new graph data augmenter under open-world settings with proper sparsity control, such that the augmented graph can be closer to the true data distribution.

    In light of the vast development of large language models (LLMs), we propose a novel framework, namely \textbf{DemoGraph}, to perform contextual graph data augmentation with a generative pretrained LLM. 
Our {contributions} can be summarized as 
    (1) 
    We introduce a black-box method which leverages extensive knowledge from LLM to perform graph data augmentation without access to model weights or source codes.
    This is particularly realistic when most LLMs are provided in close-source commercial APIs, enabling the democratization of LLM-based methods. 
        We adopt latent KGs to capture the structural interactions from the text outputs, as well as a compatible data structure for graph data.
    (2) We design a dynamic merging strategy to stochastically integrate the LLM-generated KGs into the raw graph data during the network training, which guides the optimization trajectory with contextual knowledge. 
    (3) To tackle the sparsity induced by generated KGs, we design a granularity-aware prompting strategy to control the sparsity while maximizing the utility of domain knowledge. Also, we leverage a sequential prompting with instruction fine-tuning strategy to incentivize the LLM to generate the most relevant concepts to the context, and hence high-quality KGs.
    (4) Extensive experiments on various graph learning tasks validate the effectiveness of our method over existing graph data augmentation methods. 
    (5) Our method demonstrates high scalability across datasets ranging from small to large-scale, consistently delivering satisfactory performance.
      Notably, our approach excels in scenarios involving electronic health records (EHRs), where our method maximizes the utilization of contextual information, and leads to enhanced predictive performance and interpretability. 
      %

\section{Related Works}

\para{Graph Neural Networks (GNNs).}
GNNs are gaining significant success in many problem domains \cite{kojima2020kgcn, hu2020HGT, liu2020hsgnn, chan2023SUMSHINE2, simonovsky2018graphvae, wu2020graph}.
They learn node representation by aggregating information from the neighboring nodes on the graph topology. 
Most of the existing GNN architectures are on homogeneous graphs \citep{kipf2016gcn, velivckovic2017GAT, xu2018GIN, yun2019GTN}.
There are also GNN architectures  operating on heterogeneous graphs to learn its enriched structural information and complex relations
\citep{wang2019HAN, hu2020HGT,huang2020dahgt, yang2020MuSDAC, schlichtkrull2018RGCN} .
However, due to limited samples, it is difficult to approximate the true data distribution, especially in the graph domain.
Hence, an effective graph data augmentation algorithm is needed to boost the performance of GNNs.

\para{Graph Data Augmentation.}
Graph data augmentation (GDA) aims to enhance the utility of the input graph data and produce graph samples close to the true data distribution to alleviate the finite sample bias \cite{ding2022GDAsurvey}.
Most of the existing works focus on perturbating the graph structures or node features/labels to achieve augmentation,
such as node dropping \cite{feng2020dropnode}, edge perturbation \citep{rong2019dropedge, velivckovic2018DGinfomax}, graph rewriting \cite{wang2020nodeaug, yang2019topology, franceschi2019learning}, graph sampling \cite{hamilton2017graphSAGE, hamilton2017inductive, qiu2020gcc}, graph diffusion \cite{topping2021understanding, zheng2020robust, qiu2020gcc, park2021metropolis} or pseudo-labelling \cite{zhang2017mixup}.
There are also works that adopt a learnable graph data augmenter and design specific losses for training \cite{you2020does, wu2020graph, liu2022local, suresh2021AD-GCL, li2018deeper, park2022graph}.
However, these methods mainly focus on the graph structures without considering the contextual information or introducing open-world knowledge.
Recent works \cite{jiang2023graphcare, he2023harnessing, west2021symbolic, zhang2022greaselm, wei2024llmrec, tang2024graphgpt} on LLM-based GDA have achieved promising improvements. However, current LLM-based methods are mostly white-box which require access to the weights or latent features from the LLMs. It is computationally inefficient and impractical, as SOTA LLMs are costly for large-scale experiments and often closed-source.
Moreover, these methods mostly focus on node-level context and neglect the higher-order graph structures.
Hence, a black-box LLM-based GDA framework with awareness of higher-level graph structure is needed to address these limitations.
%

\para{Graph Learning in Healthcare.}
Knowledge distillation from massive EHRs has been a popular topic in healthcare informatics. 
To address the longitudinal features in the EHR data, several early works \citep{ma2017dipole, ma2020concare, ma2020adacare} attempted to learn the EHR features with recurrent neural networks.
Since the EHR data represent relational information between entities (e.g., patients make visits), graphical models turn out to be an ideal approach for representing the EHR data \citep{choi2017gram, choi2018mime}.
GRAM \citep{choi2017gram} is a well-known method that learns robust medical code representations by adopting a graph-based attention mechanism.
However, a critical gap remains in these methods: they do not fully incorporate the rich contextual information available in EHR data \cite{hsu2012context, del2013disseminating}. This oversight can lead to a lack of nuanced understanding of patient data, impacting the accuracy and applicability of the insights derived \cite{evans2016electronic}. 
Furthermore, there is a notable absence of effective regularization mechanisms for adjusting to the inherent noise in EHR data, which is cluttered with irrelevant or redundant information.

\begin{figure*}
    \centering
    \includegraphics[width=\textwidth]{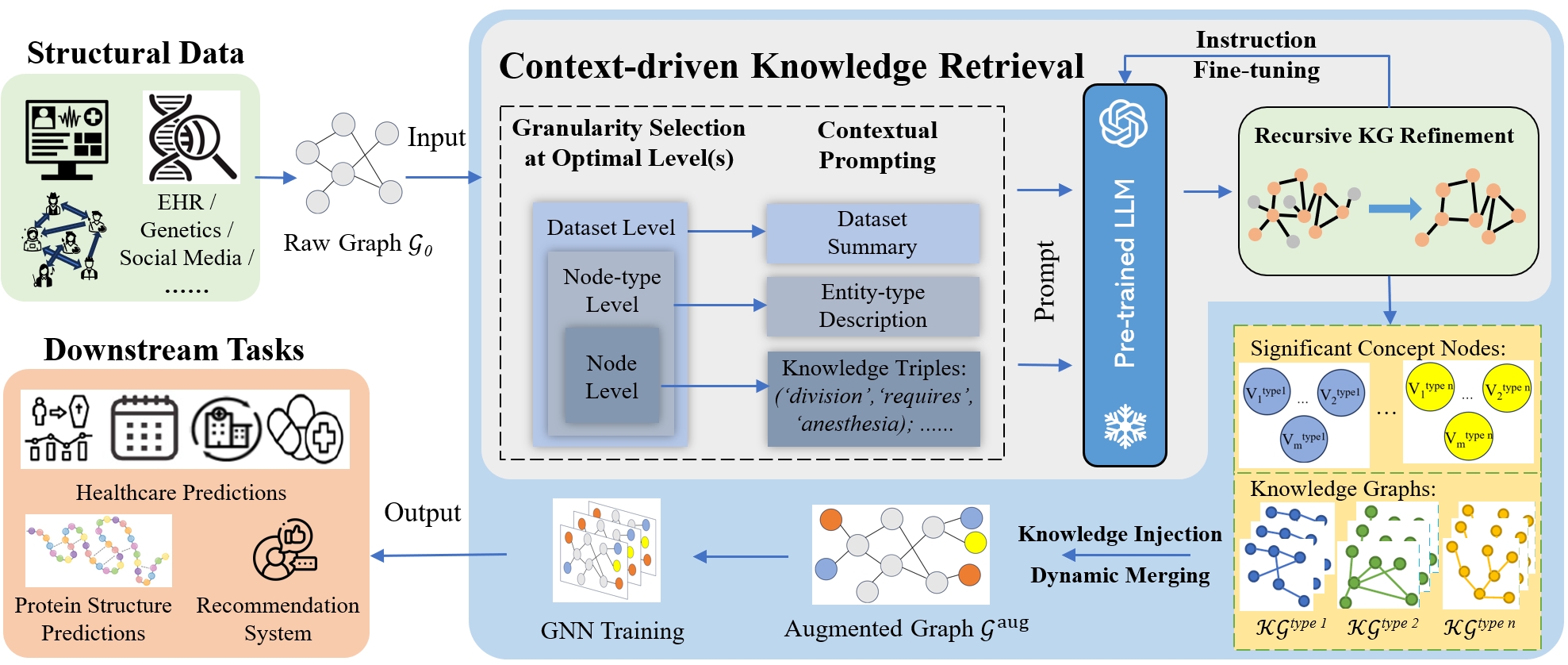}
    \caption{\textbf{Overview of our proposed DemoGraph framework.} \textnormal{Given a dataset, we first construct a 
    graph \( \mathcal{G}_0 \) to highlight the relational information, and
    then perform context-driven knowledge retrieval by utilizing the original dataset and a frozen generative pre-trained LLM. We conduct contextual, adaptive, sparsity-controllable and granularity-aware prompt learning on the LLM, thus obtaining either concept-specific KGs or important extra concept nodes at different levels after refinement. For the original graph \( \mathcal{G}_0 \), we perform graph data augmentation with the domain-knowledge injection procedure. 
    We train a GNN model on the augmented graph \( \mathcal{G}^\text{aug} \), thus our framework is able to handle a wide range of downstream tasks across various domains depending on the original datasets. }}
    \label{fig: framework}
\end{figure*} 

\section{Preliminaries}

\para{Graphs.}
A graph $\mathcal{G}$ is a collection of vertices $\mathcal{V}$ and edges $\mathcal{E}$, typically represented as $\mathcal{G} = (\mathcal{V}, \mathcal{E})$. Each edge $e \in \mathcal{E}$ is an ordered or unordered pair of vertices representing the connection between them. 
In the context of graph neural networks, each vertex $v_i$ is often associated with a feature vector $x_i$ in the feature space $\mathcal{X}$. 
A knowledge graph (KG) is a specialized type of graph 
denoted as $\mathcal{KG} = (\mathcal{V}, \mathcal{E}, \mathcal{R})$, where $\mathcal{R}$ is a set of relation types. 
A KG can be constructed from a set of triples $\mathcal{T} = \{(h_i, r_i, t_i)\}_{i = 1}^{|\mathcal{T}|}$ where $h_i, t_i$, and  $r_i$ are the $i$-th head and tail nodes respectively, and $r_i$ is the relation type for the $i$-th triple.

\para{Graph Data Augmentation (GDA).}
Given $\mathcal{G}$ = $(\mathcal{V}, \mathcal{E})$, GDA aims to derive an augmented graph $\mathcal{G}^{\text{aug}}$ = $(\mathcal{V}^{\text{aug}}, \mathcal{E}^{\text{aug}})$, where $\mathcal{V}^{\text{aug}}$ and $\mathcal{E}^{\text{aug}}$ represent the augmented set of nodes and edges, respectively. The augmentation process should preserve or enhance the inherent structure and properties of $\mathcal{G}$, while facilitating improved performance of a GNN (denoted as $\mathcal{M}$) on downstream tasks.

\section{Methodology}
Our proposed framework consists of two main modules: a knowledge graph construction module with leveraging knowledge from LLMs, and a graph data augmentation module with dynamic knowledge injection. Figure \ref{fig: framework} and Algorithm \ref{alg: workflow} provide an overview of the workflow of our framework.
\begin{figure}
    \begin{minipage}{0.5\textwidth}
    \begin{algorithm}[H]
    \begin{algorithmic}[1]
    \State \textbf{Input:} Original graph \( \mathcal{G}_0 = (\mathcal{V}_0, \mathcal{E}_0)\) with randomly-initialized node features \( \{x_i, \forall i \in \mathcal{V}\} \), granularity level $s$, number of KGs generated $K$ (per step), ground truth labels $y$.
    \State \textbf{Output:} Augmented graph \( \mathcal{G}^{\text{aug}} \), trained GNN model $\mathcal{M}$.
    \State Initialize \( \mathcal{G}^{\text{aug}} = \mathcal{G}_0 \)
    \For{each epoch}
        \State \( \mathcal{V}^{\mathcal{KG}}\leftarrow  \)
        Get concept nodes as augmentation entities,
        \State  $\{\mathcal{KG}\}_{i=1}^K$ $\leftarrow$  Load KGs from $\mathcal{V}^{\mathcal{KG}}$,
        \State $\{\mathcal{KG}\}_{i=1}^K$ $\leftarrow$ Perform instruction fine-tuning with customized sparsity control on $\{\mathcal{KG}\}_{i=1}^K$,  
        \State \( \mathcal{G}^{\text{aug}} \leftarrow \text{merge\_KG}(\{\mathcal{KG}\}_{i=1}^K, \mathcal{G}^{\text{aug}}) \),
        \State Update node indices for all node types in \( \mathcal{G}^{\text{aug}} \),
        \State Get prediction from the GNN \( \hat{y} =  \mathcal{M}(\mathcal{G}^{\text{aug}}) \),
        \State Compute training loss \( \mathcal{L}(\hat{y}, y) \),
        \State Backpropagate $\mathcal{L}$ to $\mathcal{M}$
    \EndFor
    \State \textbf{return} Trained GNN $\mathcal{M}$
    \end{algorithmic}
    \caption{The training workflow of our graph data augmentation method.}
    \label{alg: workflow}
    \end{algorithm}
    \end{minipage}
\end{figure}
\subsection{Context-Driven Knowledge Retrieval}

\para{General Prompting Strategy.}
The cornerstone of our framework is the construction of KGs using LLMs. 
The context-aware KGs serve as enriched contextual domain knowledge that augments the original graph $\mathcal{G}_0$ towards the true representation $\mathcal{G}_t$.
The KG construction is facilitated through a prompting mechanism that steers the LLM toward generating subgraphs focused on specific concepts.
The generation process in general can be formulated as 
$
    \mathcal{T} \leftarrow \text{LLM}(\text{prompt}) ,  
$
where $\mathcal{T} = \{(h_i, r_i, t_i)\}_{i = 1}^{|\mathcal{T}|}$ represents the set of triples indicating the relationships between generated concepts. 
A knowledge graph $\mathcal{KG}$ can then be constructed from $\mathcal{T}$.  
We design modularized prompts (with placeholders for the descriptions) that are based on all the available information (e.g., the summary of datasets, task descriptions) of the working graph dataset, such that context knowledge can be maximally utilized. 
One example of the prompting design on the EHR context is:

\begin{lstlisting}[] 
Start with the following prompt on a given medical concept (such as health condition/treatment procedure/drug) and generate an extensive array of associated connections based on your domain knowledge. These connections should help improve prediction tasks in healthcare, e.g. drug recommendation, mortality prediction, length of stay and readmission prediction.
Format each association as [ENTITY 1, RELATIONSHIP, ENTITY 2], ensuring the sequence reflects the direction of the relationship. Both ENTITY 1 and ENTITY 2 are to be nouns. Elements within [ENTITY 1, RELATIONSHIP, ENTITY 2] must be definitive and succinct.
Approach in both breadth and depth. Continue expanding [ENTITY 1, RELATIONSHIP, ENTITY 2] combinations until reaching a total of 100.
{example}
prompt: {descriptions}
updates:
\end{lstlisting}
where the variables as placeholders are inside $\{\}$ --- \{example\} provides an exemplar triple format, \{descriptions\} provides the contextual information, and ``updates:" prompts the LLM to finish the paragraph.
This prompt initially instructs the LLM to identify and generate concept entities $\mathcal{V}^{\mathcal{KG}}$ and their interrelations $\mathcal{E}^{\mathcal{KG}}$ driven by the descriptions (e.g., on the dataset or entity) and oriented to the target tasks.
Subsequently, the LLM regularizes these relationships into standardized triple formats.
Finally, the above prompt expands this structured information both in width and depth, digging into more meaningful and nested relationships, until a pre-defined number of triples is reached.
We also prompt example triples to regularize the output formats of $\mathcal{T}$.
This multi-step process ensures that the KG is both information-rich and aligned with domain-specific objectives.
Notably, this paradigm utilizing placeholders avoids manual prompt customization, thereby reducing human labor costs.
%

\para{Granularity-Aware Prompting for Sparsity Control.}
Naively utilizing the prompting strategy in the previous section would mostly lead to a sparse KG, where data points are unevenly distributed with many gaps or missing links. Hence, we propose a multi-layer augmentation strategy that determines a granularity level prior to generation, such that the sparsity of the KG can be controlled. 

Granularity refers to the data scale of detail in the augmentation process, ranging from coarse-grained dataset-level to fine-grained node-level information.
Based on the availability of information in the working dataset, we define $s$ as the sparsity level parameter ($s$ increases as the data are more fine-grained), and separate the prompting strategy into three granularity levels, $s_0 < s_1 < s_2 $, as follows: 
\begin{itemize}
    \item \textbf{Dataset-level Augmentation ($s = s_0$).} At the dataset level, our objective is to identify and propagate overarching themes and concepts that are broadly relevant across the dataset.
    This macro approach involves curating concepts and triples that reflect high-level semantics and dependencies.
    This is the most fundamental form of our method since dataset-level information is always available.
    
    \item \textbf{Type-level Augmentation ($s = s_1$).} Another common scenario is that we have node type level information (e.g., class labels in texts for classification).
We distill the most salient concepts and relationships pertinent to each class or node type. 
By doing so, we gain an in-depth understanding of the node categories, fleshing out their characteristics and the interconnections within them.
A node-type level prompting example on the Cora dataset (7 classes) is provided in the appendix. 

\item \textbf{Node-level Augmentation ($s = s_2$).} In some scenarios (e.g., EHR datasets), we have the finest information (e.g., text description) on each node (or medical entity). 
At this juncture, we aim to enrich individual nodes with highly relevant and specific concepts that are crucial for the particular tasks. 
This targeted augmentation ensures that nodes are imbued with unique attributes that can drive predictive tasks more effectively.

\end{itemize}

\begin{figure}
    \centering
    \includegraphics[width=0.5\textwidth]{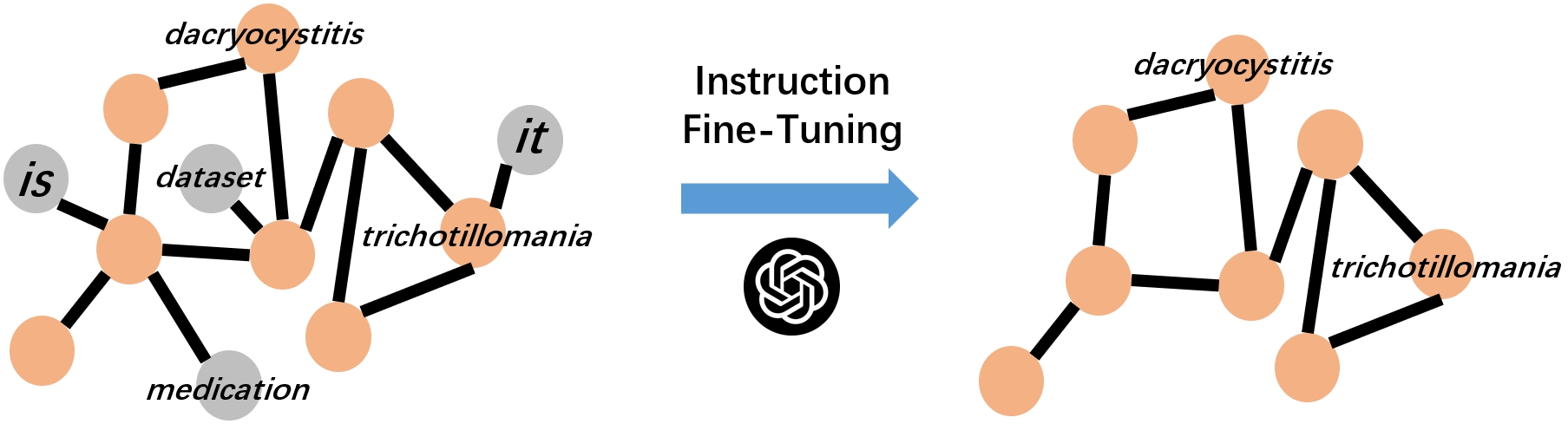}
    \caption{Concept pruning via instruction fine-tuning, where trivial concepts can be pruned by re-prompting the coarse set of concepts to the LLM.}
    \label{fig:IFT}
\end{figure}

\para{Concept Pruning via Instruction Fine-tuning.}
Due to the high complexity of given tasks, LLM's one-time retrieval of KGs may contain low-entropy (i.e., uninformative) concepts (e.g., \textit{is}, \textit{dataset}, or \textit{disease}). We thus instruct LLMs to go through a chain-of-thought process to do multi-stage reasoning and self-improve the quality of KGs.
Figure \ref{fig:IFT} illustrates our concept prompting procedure via instruction fine-tuning.
Given the initial generated $\mathcal{KG}$, we refine it by recursively calling the LLM and pruning less relevant nodes and edges, while ensuring that a predefined percentage of the concepts are directly derived from the original dataset.
A template for this instruction fine-tuning (IFT) process is given below (we use EHR as an illustrative example). After this procedure, a set of important concept nodes $\mathcal{V}^{\mathcal{KG}}$ is then output for triple construction and KG generation.
\begin{lstlisting}[]
Given the list of triples augmented with MIMIC-III dataset, I want to select '{number_of_concepts}' most important triples from the list. The importance of a triple is based on your knowledge and inference on how it will help improve prediction tasks in healthcare, e.g. drug recommendation, mortality prediction, length of stay, readmission prediction. If you think a triple is important, please keep it. Otherwise, please remove it. You can also add triples from your background knowledge.
triples: {triples}
updates:
\end{lstlisting}

\subsection{Augmentation with Generated KGs}

\para{Dynamic Graph Merging.}
Given a constructed $\mathcal{KG}$ from $\mathcal{T}$ on a sparsity level $s$, we design a dynamic merging schema to merge $\mathcal{KG}$ into $\mathcal{G}_0$. 
This allows the model to see more augmented samples $\mathcal{G}^{\text{aug}}$ as a different merged graph is obtained in each optimization step. 
For each concept node $v_c \in \mathcal{V}^{\mathcal{KG}}$ in  $\mathcal{KG}$, we select a subset of nodes $\mathcal{V}_0^s = \{z | z \in \mathcal{V}_0 \}_{i=0}^{n_c} \subseteq \mathcal{V}_0$, where $n_c$ is the predetermined number of edges per concept node.
We connect the concept nodes and the selected nodes from $\mathcal{V}_0^s$ to obtain an edge set
\begin{align*}
    \mathcal{E}^{\text{conn}} = \{ (v_c , z) | \forall v_c \in \mathcal{V}^{\mathcal{KG}}, z \in \mathcal{V}_0^s\}.
\end{align*}
After that, the augmented graph $\mathcal{G}^{\text{aug}} =$ ($\mathcal{V}^{\text{aug}}, \mathcal{E}^{\text{aug}}$) can be obtained by joining the edge sets and node sets, i.e., $\mathcal{E}^{\text{aug}} = \mathcal{E}^{\text{conn}} \cup \mathcal{E}_0 \cup \mathcal{E}^{\mathcal{KG}} $ and $\mathcal{V}^{\text{aug}} = \mathcal{V}_0 \cup \mathcal{V}^{\mathcal{KG}}$.

This dynamic merging is not a one-off operation but an iterative process. Each training epoch sees the refreshment of KGs based on the model's current state, thereby keeping the graph data dynamic and contextually rich. 
As the model training proceeds, it continually refines the edge weights and node features based on the newly incorporated KGs. 
This iterative update ensures that the model does not overfit and generalizes well on unseen data.

Due to the computation limitations, the number of LLM inferences is limited.
Therefore, we precompute $\mathcal{KG}$ offline and merge it with $\mathcal{G}_0$ stochastically during training.
Under sufficient computational conditions, the dynamic merging schema allows for online prompting where an up-to-date $\mathcal{KG}$ can be generated after every optimization step.
On the other hand, the LLM can also be fine-tuned online with task-specific losses.
This allows for more context-related KG generations and hence improved data augmentation performance. 
It also enables the potential for training open-world GNN models.

\para{Training Paradigm.}
We use GNN to predict the labels with the augmented graph as the input,
$
    \hat{y} = \mathcal{M}(\mathcal{G}^{\rm aug}).
$ We benchmark with different choices of $\mathcal{M}$: graph convolutional network (GCN) \cite{kipf2016gcn}, graph attention network (GAT) \cite{velivckovic2017GAT}, GraphSAGE \cite{hamilton2017graphSAGE}, and graph isomorphism network (GIN) (detailed formulations and descriptions of GNNs in appendix).
We compute the loss for backpropagation with the predictive labels.
For instance, in a multi-class classification task, we adopt the cross-entropy loss,
$
    L_{\text{ce}} = -\frac{1}{N} \sum_{i=1}^{N} \sum_{c=1}^{C} y_{i,c} \log(\text{softmax}(z_{i,c})),
$
where \( y_{i, c} \) is the ground truth label for patient \( i \) and class $c$, \( N \) is the number of observations, \( C \) is the number of classes, and \( z_{i,c} \) is logits obtained from the model.

\subsection{Adaptability to Other Graph Datasets}

Since EHR contains enriched contextual information that allows for flexible prompting design, we use the EHR dataset to illustrate our prompting strategy. 
However, our prompting strategy is adaptable to other graph datasets, as the placeholders in the modularized prompts can be replaced by information on the target datasets. 
We can also incrementally enlarge the KG such that knowledge from the existing domain can be leveraged to the target domain.
We employ a highly-adaptive customization strategy that tailors the prompt structure based on the specific dataset in use. 
This strategy includes understanding the data's content and structure and then adjusting the prompts to ensure the generated KGs are optimally suited for the data in question.

\begin{table}[]
    \centering
    \captionof{table}{Node classification performance (in common metrics of existing literature) on generic graph datasets with different GNN architectures. Standard deviations are shown in brackets.}
    \centering
    \scalebox{0.65}{
    \begin{tabular}{>{\centering\arraybackslash}m{0.8cm}|l|c|c|c|c}
    \toprule
      &  & \textbf{PPI} & \textbf{Actor} & \textbf{Cora} & \textbf{Citeseer} \\
     \textbf{GNN Archi.} & \textbf{Augmenter} & \textbf{Micro-F1} & \textbf{Accuracy}  & \textbf{Accuracy} & \textbf{Accuracy} \\
    \hline
    \multirow{6}{1.2cm}{\textbf{Graph\\SAGE}} & None & 60.0 (2.7) & 36.7 (1.8) & 81.0 (3.3) & 70.9 (2.0) \\
    & DropNode \citep{feng2020dropnode} & 61.5 (2.6) & 36.8 (1.5) & 80.6 (3.2) & 70.1 (2.7) \\
    & DropEdge \citep{rong2020dropedge} & 63.2 (3.1) & 36.8 (2.9) & 80.4 (2.8) & 71.2 (3.2) \\
    & RandomwalkPE \citep{dwivedi2021randomwalkPE} & 63.1 (2.7)& 37.7 (2.7) &81.2 (3.1) & 70.8 (2.6) \\
    & LaplacianPE \citep{dwivedi2023laplacianPE} & 63.5 (3.1) & 36.7 (2.1) &80.9 (2.2) & 70.7 (2.5) \\
    & \textbf{DemoGraph (Ours)} & \textbf{93.6 (2.3)} & \textbf{37.9 (1.6)} & \textbf{83.3 (1.2)} & \textbf{72.6 (2.0)} \\ \hline
    \multirow{6}{*}{\textbf{GAT}} &None & 97.1 (3.0) & 30.3 (2.7) & 82.1 (4.3) & 72.1 (3.7) \\
    & DropNode  \citep{feng2020dropnode}& 94.0 (3.4) & 31.3 (2.2) & 80.7 (3.7) & 71.9 (3.2) \\
    & DropEdge \citep{rong2020dropedge} & 85.1 (3.0) & 31.2 (3.0) & 78.9 (3.9) & 69.1 (3.9) \\
    & RandomWalkPE \citep{dwivedi2021randomwalkPE} & 90.8 (3.6) & 31.4 (2.5) & 81.2 (3.2) & 71.9 (3.2) \\
    & LaplacianPE \citep{dwivedi2023laplacianPE} & 90.7 (2.7) & 30.9 (2.9) & 81.4 (2.4) & 71.8 (2.7) \\
    & \textbf{DemoGraph (Ours)} & \textbf{97.2 (3.4)} & \textbf{32.2 (2.3)} & \textbf{83.6 (2.0)} & \textbf{73.1 (2.2)} \\ \hline
    \multirow{6}{*}{\textbf{GCN}} & None & 53.2 (2.4) & 29.8 (2.1) & 81.0 (2.7) & 69.4 (2.0) \\
    & DropNode \citep{feng2020dropnode} & 58.9 (1.9) & 28.7 (2.5) & 78.9 (2.6) & 70.5 (2.0)  \\
    & DropEdge \citep{rong2020dropedge} & 54.8 (4.1) & 28.9 (3.4) & 82.4 (3.5) & 71.3 (3.2)\\
    & RandomWalkPE \citep{dwivedi2021randomwalkPE} & 59.0 (1.6) & 29.8 (2.9) & 80.0 (2.9) & 71.6 (2.2)  \\ 
    & LaplacianPE \citep{dwivedi2023laplacianPE} & 59.3 (1.6) & 29.6 (2.2) & 80.0 (1.9) & 71.1 (2.1) \\
     & \textbf{DemoGraph (Ours)} & \textbf{60.3 (1.2)} & \textbf{32.4 (2.3)} &\textbf{82.9 (1.0)} & \textbf{73.1 (1.1)}\\
    \hline
    \multirow{6}{*}{\textbf{GIN}} & None & 70.3 (2.8) & 31.9 (2.0) & 81.6 (2.0) & 70.9 (3.7) \\
    & DropNode \citep{feng2020dropnode} & 75.2 (3.1) & 32.4 (2.2) & 78.5 (4.1) & 70.6 (4.0) \\
    & DropEdge \citep{rong2020dropedge} & 78.3 (3.7) & 32.7 (2.8) & 81.8 (4.4) & 71.5 (3.9) \\
    & RandomWalkPE \citep{dwivedi2021randomwalkPE} & 76.2 (3.5) &33.1 (2.5) &80.9 (2.7) & 71.1 (3.8)\\
    & LaplacianPE \citep{dwivedi2023laplacianPE} & 74.5 (2.9) &32.9 (2.4) &81.9 (2.7)& 71.4 (3.6) \\ 
    & \textbf{DemoGraph (Ours)} & \textbf{79.2 (2.8)} & \textbf{34.8 (2.2)} & \textbf{82.3(4.5)} & \textbf{72.9 (3.9)} \\ \bottomrule
    \end{tabular}}
        \label{tab: graph_aug_generic}
\end{table}

\begin{table}[]
    \centering
    \captionof{table}{Performance [\%] of DemoGraph on the OGBN-arxiv and OGBN-products datasets.}
    \centering
    \scalebox{0.7}{
        \begin{tabular}{>{\centering\arraybackslash}m{0.8cm}|c|c|c}
        \toprule
        \textbf{GNN Archi.} & \textbf{Augmenter} & \multicolumn{2}{c}{\textbf{Accuracy}} \\ \cline{3-4}
                            &                    & \textbf{OGBN-products} & \textbf{OGBN-arxiv} \\ \hline
        \textbf{Graph } & DropNode & 54.22 (0.31) & 58.42 (0.20) \\
                        \textbf{SAGE}   & DropEdge & 55.23 (0.32) & 54.83 (0.19) \\
                           & RandomWalkPE & OOM & OOM \\
                           & LaplacianPE & OOM & OOM \\
                           & GraphGPT-std & N/A & 62.58 \\
                           & LLM* & 74.40 (0.23) & 73.56 (0.06) \\
                           & TAPE & 81.37 (0.43) & 76.72 (0.07) \\
                           & \textbf{DemoGraph (Ours)} & \textbf{84.22 (0.27)} & \textbf{76.84 (0.17)} \\ \hline
        \textbf{GAT}       & DropNode & 55.43 (0.34) & 57.36 (0.25) \\
                           & DropEdge & 53.36 (0.37) & 58.26 (0.21) \\
                           & RandomWalkPE & OOM & OOM \\
                           & LaplacianPE & OOM & OOM \\
                           & GraphGPT-std & N/A & 62.58 \\
                           & LLM* & 74.40 (0.23) & 73.56 (0.06) \\
                           & TAPE & 82.34 (0.36) & \textbf{77.50 (0.12)} \\
                           & \textbf{DemoGraph (Ours)} & \textbf{84.00 (0.32)} & \textbf{77.18 (0.22)} \\ \hline
        \textbf{GCN}       & DropNode & 56.94 (0.45) & 58.57 (0.42) \\
                           & DropEdge & 54.62 (0.47) & 58.15 (0.43) \\
                           & RandomWalkPE & OOM & OOM \\
                           & LaplacianPE & OOM & OOM \\
                           & GraphGPT-std & N/A & 62.58 \\
                           & GraphGPT-stage2 & N/A & 75.11 \\
                           & 3-HiGCN & N/A & \textbf{76.41 (0.53)} \\
                           & LLM* & 74.40 (0.23) & 73.56 (0.06) \\
                           & TAPE & 79.96 (0.41) & 75.20 (0.03) \\
                           & \textbf{DemoGraph (Ours)} & \textbf{82.86 (0.42)} & \textbf{76.05 (0.23)} \\ \hline
        \end{tabular}}
    {\scriptsize \begin{spacing}{0.5}
    OOM: out-of-memory.
    LLM: Using zero-shot ChatGPT with the same prompts of TAPE as the approach, denoted as LLM.
    \end{spacing}}
    \label{tab: OGBN_results_combined}
\end{table}

\begin{table*}[]
    \caption{Performance of drug recommendation, length of stay, mortality and readmission prediction on MIMIC-III [\%]. Standard deviations are shown in brackets.}
    \centering
    \scalebox{0.85}{
        \begin{tabular}{l|cc|cc|cc|cccc p{1.3cm}}
        \toprule
        \multicolumn{1}{c|}{}& \multicolumn{2}{c|}{\textbf{Drug Recommendation}} & \multicolumn{2}{c|}{\textbf{Length of Stay}} & \multicolumn{2}{c|}{\textbf{Mortality}} & \multicolumn{2}{c}{\textbf{Readmission}} \\
        \multicolumn{1}{c|}{\textbf{Model}} &
        \multicolumn{1}{c}{\textbf{AUROC}}& 
        \multicolumn{1}{c|}{\textbf{AUPR}} &
        \multicolumn{1}{c}{\textbf{AUROC}}& 
        \multicolumn{1}{c|}{\textbf{Acc.}} &
        \multicolumn{1}{c}{\textbf{AUROC}}& 
        \multicolumn{1}{c|}{\textbf{AUPR}} &
        \multicolumn{1}{c}{\textbf{AUROC}}& 
        \multicolumn{1}{c}{\textbf{AUPR}} 
        \\ \hline
        GRU  & 96.38 (0.1) & 64.75 (0.2) & 80.32 (0.2) & 42.14 (0.6) & 61.09 (0.7) & 9.65 (1.5) & 65.58 (1.1) & 68.57 (1.6) \\ 
        Transformer & 95.87 (0.0) & 60.19 (0.1) & 79.31 (0.8) & 41.68 (0.7) & 57.20 (1.3) & 10.10 (0.9) & 63.75 (0.5) & 68.92 (0.1)  \\
        Deepr 
        & 96.09 (0.0) & 62.48 (0.1) & 78.02 (0.4) & 39.31 (1.2) & 60.80 (0.4) & 13.20 (1.1) & 66.50 (0.4)  & 68.80 (0.9)  \\
        GRAM
        & 94.20 (0.0) & 76.70 (0.1) & 78.02 (0.4) & 39.31 (1.2) & 60.40 (0.9) & 11.40 (0.7) & 64.30 (0.4) & 67.20 (0.8) \\
        Concare
        & 95.78 (0.1) & 61.67 (0.3) & 80.27 (0.3) & 42.04 (0.6) & 61.98 (1.8) & 9.67 (1.5) & 65.28 (1.1) & 66.67 (1.9)  \\
        Dr. Agent
        & 96.41 (0.1) & 64.16 (0.5) & 79.45 (0.6) & 41.40 (0.5) & 57.52 (0.4) & 9.66 (0.8)  & 64.86 (2.6) & 67.41 (1.0)\\ 
        AdaCare
        & 95.86 (0.0) & 60.76 (0.0)  & 78.73 (0.4) & 40.70 (0.8) & 58.40 (1.4) & 11.10 (0.4)  & 65.70 (0.3) & 68.60 (0.6)  \\ 
        StageNet 
        & 96.05 (0.0) & 62.43 (2.4) & 77.94 (0.2) & 40.70 (0.8) & 61.50 (0.7) & 12.40 (0.3)  & 66.70 (0.4) & 69.30 (0.6) \\
        GRASP 
        & 96.01 (0.1) & 62.53 (0.3) & 78.97 (0.4) & 40.66 (0.3) & 59.20 (1.4) & 9.90 (1.1)  & 66.30 (0.6) & 69.20 (0.4) \\ \hline
        DropNode & 97.60 (0.2) & 81.41 (0.1) & 81.10 (0.5) & 41.81 (1.1) & 58.06 (0.9) & 9.46 (1.7) & 64.48 (0.8) & 67.75 (0.4) \\ 
        DropEdge & 95.61 (0.1) & 72.32 (0.3) & 78.41 (0.3) & 39.98 (0.8) & 57.85 (0.8) & 10.34 (1.5) & 62.11 (0.6) & 67.46 (0.5) \\ 
        RandomWalkPE & 94.89 (0.1) & 63.86 (0.2) & 78.01 (0.4) & 39.47 (0.9) & 57.15 (1.2) & 9.76 (0.9) & 66.20 (0.7) & 59.58 (0.6) \\ 
        LaplacianPE & 95.26 (0.2) & 69.34 (0.3) & 78.22 (0.3) & 40.02 (0.9) & 57.65 (1.1) & 10.05 (1.2) & 65.71 (0.6) & 63.43 (0.8)\\ 
        GraphCare 
        & 95.00 (0.0) & 78.50 (0.2) & 79.40 (0.3) & 41.90 (0.2) & 66.60 (1.1) & 14.30 (0.8) & 68.10 (0.6) & 71.50 (0.7)
        \\ \hline         
        \multicolumn{1}{c|}{\textbf{DemoGraph (Ours)}} & \textbf{98.54 (0.2)} & \textbf{83.89 (0.1)} &\textbf{82.68 (0.2)} & \textbf{45.28 (1.0)} & \textbf{67.79 (0.6)} & \textbf{16.09 (1.6)} & \textbf{68.97 (0.4)} & \textbf{73.92 (0.4)} \\ 
        \bottomrule
        \end{tabular}}
    \label{tab: results_MIMIC3}
\end{table*}

\section{Experiments}

\subsection{Experimental Settings}

\para{Datasets and Tasks. }
(1) We perform experiments on \textbf{generic} graph benchmarks (Cora, PPI, Actor, and Citeseer), where we benchmark our method on node classification tasks.
(2) We validate the scalability of DemoGraph on two \textbf{large-scale} datasets --- OGBN-products and OGBN-arxiv \cite{hu2020OGB} against additional LLM-based methods.
Table \ref{tab: datasets general graphs} and \ref{tab:OGBN_summary} provide a summary of these graph datasets from small to large-scales.
(3) Additionally, we highlight an application of our method on a \textbf{large-scale EHR} dataset --- MIMIC-III \cite{johnson2016mimic}.
It contains a publicly available dataset of 46,520 intensive care unit (ICU) patients over 11 years.
We perform four supervised tasks --- in-hospital mortality prediction (MORT), readmission prediction (READM), length of stay (LOS) prediction, and drug recommendations (DR), where MORT and READM predictions are approached as binary classification tasks, LOS prediction as a multi-class classification task, and DR as a multi-label classification task.
Since the lab events are sparse and introduce heavy noise, we exclude them when constructing the graph.
Table \ref{tab: mimic3_summary} in the appendix presents a summary of the types and counts of the entities in the MIMIC-III dataset, and the details of each task.

\para{Evaluation Metrics.} We evaluate our method with area under the receiver operating curve (AUROC), area under the precision-recall curve (AUPR), accuracy, F1-scores, and Jaccard index, applied as relevant to each task.
For robust validation of our results, we employ a five-fold cross-validation strategy in all major experiments.
More detailed information on the datasets, tasks and their loss functions, and evaluation metrics is presented in the appendix.

\subsection{Compared Methods}
We compare our method to the following graph data augmentation methods to validate the empirical performance of DemoGraph: LaplacianPE \citep{dwivedi2023laplacianPE}, RandomWalkPE \citep{dwivedi2021randomwalkPE}, DropEdge \citep{rong2020dropedge}, and  DropNode \citep{feng2020dropnode}.
For the EHR analysis benchmark, we also include additional competitors as follows: GraphCare (LLM-based) \citep{jiang2023graphcare}, GRU \citep{medsker2001RNN}, Transformer \citep{vaswani2017transformer}, GRAM \citep{choi2017gram}, StageNet \citep{gao2020stagenet}, Concare \citep{ma2020concare}, Adacare \citep{ma2020adacare}, Dr. Agent \citep{gao2020dragent}, and GRASP \citep{zhang2021grasp}.
For drug recommendation, we also include additional competitors: MICRON \citep{yang2021micron}, Safedrug \citep{yang2021safedrug}, and MoleRec \citep{yang2023molerec}.
For the large-scale OGBN datasets, additionally, we have included more advanced LLM-based baselines (i.e., GraphGPT \cite{tang2023graphgpt}, LLM, TAPE \cite{he2023harnessing} and HiGCN \cite{huang2024higher}).
We reimplemented the baseline methods, where details of the implementations and descriptions of the baseline methods can be found in the appendix.
%

\subsection{Quantitative Results}
\para{Results on Generic Graph Data.} Table~\ref{tab: graph_aug_generic} presents the node classification results of our proposal compared to existing graph data augmentation methods.
Table~\ref{tab: OGBN_results_combined} presents the results on the large-scale OGBN-products and OGBN-arxiv datasets against both traditional and LLM-based competitors.
We observe that our method achieves satisfactory performance on generic graph classification datasets, as well as large-scale datasets.
Some of the traditional GDA methods which operate on whole graphs failed to generalize to large-scale datasets (i.e., encountered out-of-memory error). 
Our method obtains a 3\% improvement on average over all comparable methods with all four GNN architectures (i.e., GCN \cite{kipf2016gcn}, GAT \cite{velivckovic2017GAT}, GIN \cite{xu2018GIN}, and GraphSAGE \cite{hamilton2017graphSAGE}).
This shows evidence that leveraging context knowledge, such as dataset summary and class label information, with LLMs can augment graph data to its true data distribution.
We also compare among the comparable methods with different GNN architectures.
We observe that our method still performs satisfactorily when different GNN architectures are used, demonstrating the robustness of our method.

\para{Results on EHR Data.} Table~\ref{tab: results_MIMIC3} presents the results of different tasks on the MIMIC-III dataset (detailed results with more evaluation metrics are presented in the appendix). 
We observe that our proposed framework outperforms alternative methods, thereby validating the effectiveness of contextual LLM augmentation and sparsity-aware instruction prompting. 
In particular, our method outperforms the competitors by 7.4\% (in accuracy) in length-of-stay prediction.
Our method can even outperform the methods specifically designed for EHR analysis, including GraphCare \cite{jiang2023graphcare}, a similar method using LLM for personalized healthcare.
We elaborate the key differences between our method and GraphCare in the appendix.
When integrating the enriched context information (e.g., clinical discharge reports, radiology reports, and lab event reports) in real-world EHR datasets, the performance on clinical task prediction can be further improved. 

\para{The Effect of Different LLM backbones.} 

In light of the importance of LLM backbones on the performance of our method, we further study the effects of LLM backbones with differnet capacities.
We performed experiments with some renowned black-box LLMs (we access these LLMs only through APIs) shown in Table \ref{tab: ablation_MIMIC3}. We observe the differences in model performances, which arise from different training methods and parameter sizes. Nevertheless, our method can maintain satisfactory performance across different LLM backbones, validating its robustness.

\begin{table}[H]
    \caption{Performance of mortality and readmission prediction on MIMIC-III [\%] with different LLM backbones. Standard deviations are shown in brackets.}
    \centering
    \resizebox{0.5\textwidth}{!}{
        \begin{tabular}{l|cc|cc}
        \toprule
        & \multicolumn{2}{c|}{\textbf{Mortality}} & \multicolumn{2}{c}{\textbf{Readmission}} \\
        \textbf{Models} &
        \textbf{AUROC} & 
        \textbf{AUPR} &
        \textbf{AUROC} & 
        \textbf{AUPR} 
        \\ \hline
        GraphCare (GPT-4, KG method) & 66.60 (1.1) & 14.30 (0.8) & 68.10 (0.6) & 71.50 (0.7) \\ \hline
        DemoGraph (LLaMA-3.1-8B) & 66.50 (0.9) & 14.70 (1.1) & 68.00 (0.7) & 71.10 (0.8) \\ 
        DemoGraph (Claude-3-Opus) & 66.90 (1.0) & 15.70 (0.8) & 69.00 (0.6) & 73.00 (0.7) \\ 
        DemoGraph (LLaMA-3.1-70B) & 67.10 (0.8) & 15.90 (1.0) & 69.20 (0.5) & 73.10 (0.6) \\ 
        DemoGraph (GPT-4, \textbf{original}) & 67.79 (0.6) & 16.09 (1.6) & 69.00 (0.4) & 73.92 (0.4) \\ 
        DemoGraph (LLaMA-3.1-405B) & 67.90 (0.7) & 16.30 (1.0) & 69.40 (0.5) & 73.80 (0.6) \\ 
        DemoGraph (Claude-3.5-Sonnet) & 68.00 (0.5) & 16.30 (0.8) & \textbf{69.60 (0.3)} & 73.90 (0.5) \\ 
        DemoGraph (GPT-4o-mini) & 68.00 (0.6) & \textbf{16.40 (0.7)} & 69.50 (0.4) & 74.00 (0.4) \\ 
        DemoGraph (GPT-4o) & \textbf{68.10 (0.5)} & 16.30 (0.8) & \textbf{69.60 (0.3)} & \textbf{74.10 (0.5)} \\ 
        \bottomrule
        \end{tabular}}
    \label{tab: ablation_MIMIC3}
\end{table}

\subsection{Qualitative Results}
\begin{figure}[h]
    \centering
        \begin{minipage}[t]{0.45\textwidth}
        \centering
        \includegraphics[width=0.48\textwidth, height=0.45\textwidth]{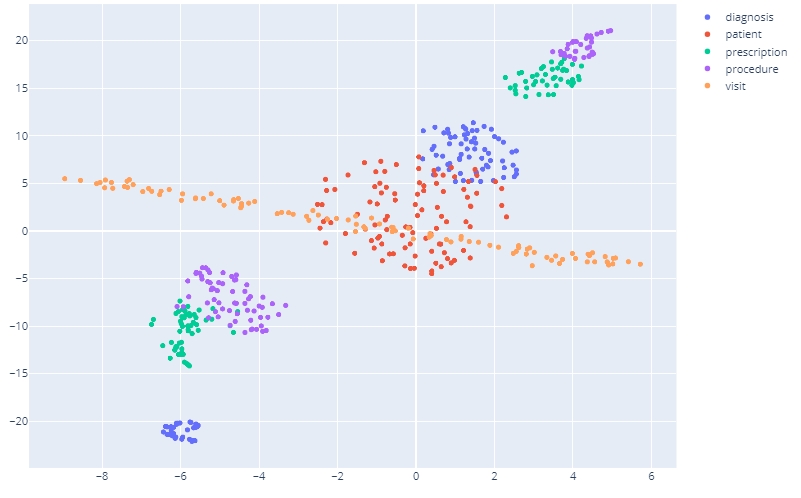}
        \includegraphics[width=0.48\textwidth, height=0.45\textwidth]{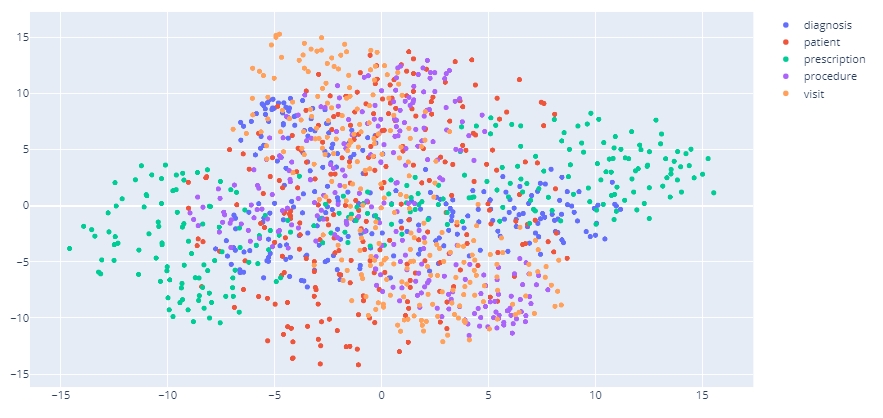}
        \caption{Visualization of the learned node embeddings w/ (left) and w/o (right) our graph data augmentation, respectively. We use MIMIC-III as the example and colour nodes differently by their entity types.}
        \label{fig: embeddings}
    \end{minipage}
    \hfill
    \begin{minipage}[t]{0.45\textwidth}
        \centering
        \includegraphics[width=0.9\linewidth]{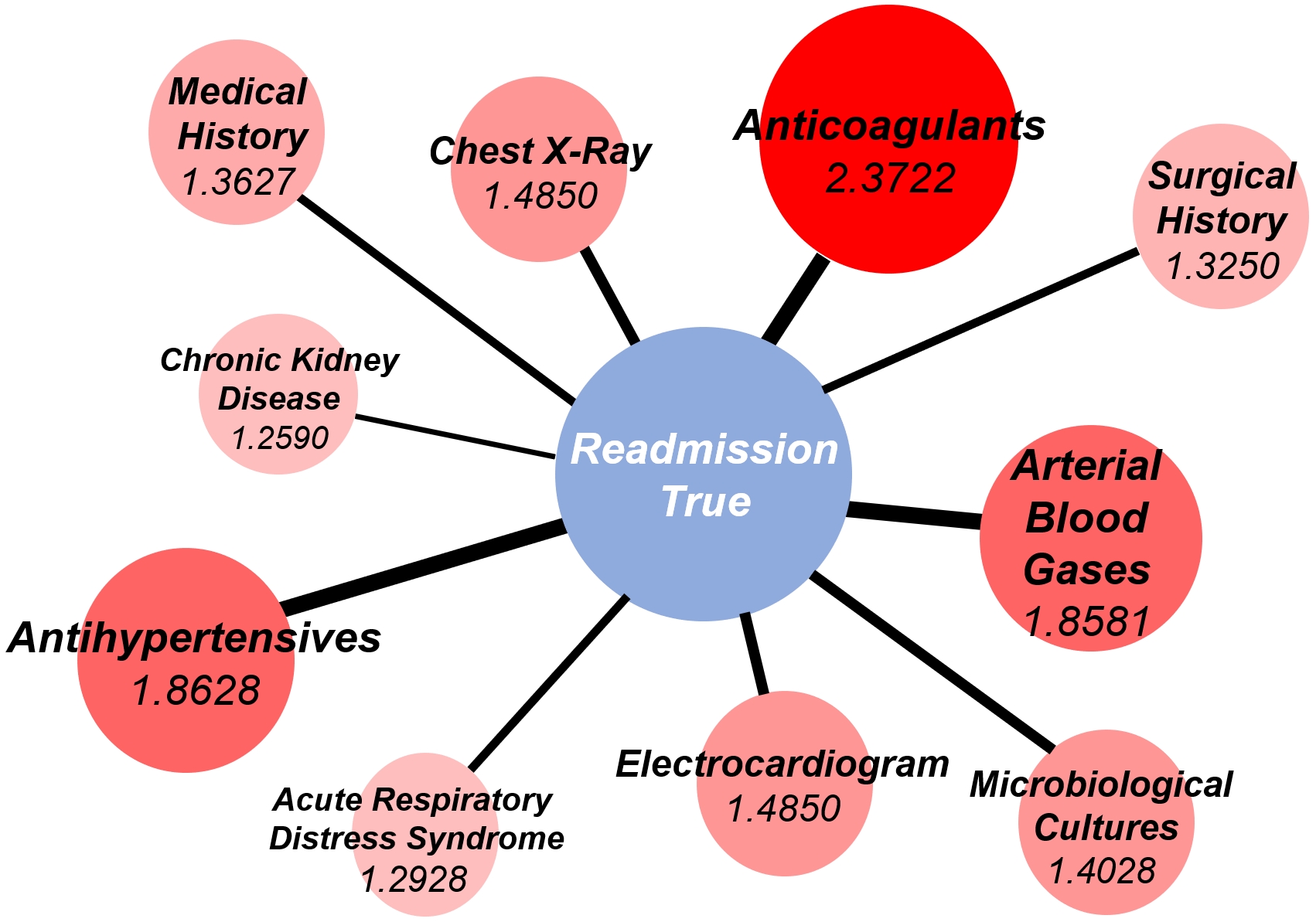}
        \caption{Visualization of the interpretability of DemoGraph: a visit node (blue) and related concept nodes (red), with attention scores visiualized in size/shade of red nodes.}
        \label{fig: interpretation}
    \end{minipage}
    \label{fig: combined}
\end{figure}

%
\para{Embedding Visualization.}
We visualize the node embeddings of each type of entity to evaluate the performance of feature representation learning.
Figure \ref{fig: embeddings} presents the TSNE plot of the embeddings generated by different methods. 
The task is readmission prediction on the MIMIC-III dataset with a GAT model.
It is observed that the embeddings with DemoGraph are grouped according to their node types, which validates that the embeddings learn the unique representation of each node type, while the embeddings without DemoGraph are noisy and do not present a clear pattern by the node type. 

\para{Network Interpretation.}
%
The incorporation of contextual learning enhances the capability of the model by enabling a nuanced understanding and interpretation of the graph data at a deeper level.
We analyze the interpretability of our model by considering a specific visit node in the MIMIC-III dataset.
As shown in Figure \ref{fig: interpretation}, the following are the top augmented corrections (i.e., with the highest attention scores) that exemplify the importance of specific clinical concepts influencing readmission prediction: Antihypertensives (2.3722), Anticoagulants (1.8628), and arterial blood gases (1.8581), where the computed attention scores are shown in brackets.
It is observed that the augmentation process can impute context-related concepts so that GAT can select the most important ones.
This provides interpretations for the predictive process.
This is especially beneficial in the clinical decision context since the enriched open-world knowledge can inspire clinicians with the embedded concepts, and enhance the understanding of patients' behaviours and the potential reasons for certain diseases.
\subsection{Ablated Analysis}

\para{The Effect of Augmented KGs. }
We study the effect of augmented KGs on downstream task performance (Table \ref{tab: w/ or w/o KG}), including three scenarios: with KG, without KG, and with a biased (or wrong) KG augmented from another dataset (i.e. PPI).
It is observed that the model performs worse than the baseline (i.e., w/o any augmentations) when the wrong context is applied, indicating a biased augmented graph.
On the other hand, improved performance is observed when a context-driven KG is applied, thus validating the effectiveness of our method.
A visualization of the effect of DemoGraph on node embeddings can also be found in Figure \ref{fig: embeddings}.

\para{The Effect of Dynamic Merging. } 
We evaluate the contribution of the dynamic merging schema, as summarized in Table \ref{tab: merging}, where static merging means that the KG are merged into $\mathcal{G}_0$ offline before training.
We observe that the performance improved on all generic graph datasets with dynamic merging, which validates the contributions of the schema. 
%

\begin{table}[H]
    \centering
    \begin{minipage}{0.45\textwidth}
        \caption{
        Performance w/ and w/o augmentation from KG, and w/ a biased KG from another dataset (i.e. PPI), repectively.
        }
        \scalebox{0.8}{
            \begin{tabular}{l|cc|cc|cc}
            \toprule
            \multicolumn{1}{l|}{} & \multicolumn{2}{c|}{\textbf{w/o KG}} & \multicolumn{2}{c|}{\textbf{w/ KG}} & \multicolumn{2}{c}{\textbf{w/ PPI KG}}\\
            \multicolumn{1}{l|}{Dataset} &
            \multicolumn{1}{c}{\textbf{Acc.}} &
            \multicolumn{1}{c|}{\textbf{F1}} &
            \multicolumn{1}{c}{\textbf{Acc.}}&
            \multicolumn{1}{c|}{\textbf{F1}} &
            \multicolumn{1}{c}{\textbf{Acc.}} &
            \multicolumn{1}{c}{\textbf{F1}}
            \\ \hline
            Cora &82.10 &81.66 &\textbf{83.60} &\textbf{83.64} & 73.70 & 73.83 \\
            Actor &30.33 &27.90 & \textbf{32.21} & \textbf{28.91} &30.19 & 27.82\\
            Citeseer & 72.10 & 69.60 &\textbf{73.10} &\textbf{72.46} &63.40 & 64.68  \\
            \bottomrule
            \end{tabular}}
        \label{tab: w/ or w/o KG}
    \end{minipage}
    \hfill
    \begin{minipage}{0.45\textwidth}
        \caption{
        Performance of node classification (using GAT) with and without dynamic merging, respectively.
        }
        \scalebox{0.64}{
            \begin{tabular}{l|cc|cc|cc|cc}
            \toprule
            &
            \multicolumn{2}{c|}{\textbf{Cora}}& \multicolumn{2}{c|}{\textbf{Actor}} & \multicolumn{2}{c|}{\textbf{PPI}} & \multicolumn{2}{c}{\textbf{Citeseer}}\\
            \multicolumn{1}{l|}{Merging} & \multicolumn{1}{c}{\textbf{Acc.}} & \multicolumn{1}{c|}{\textbf{F1}} &\multicolumn{1}{c}{\textbf{Acc.}} &\multicolumn{1}{c|}{\textbf{F1}} &\multicolumn{1}{c}{\textbf{Acc.}} &\multicolumn{1}{c|}{\textbf{F1}}
            &\multicolumn{1}{c}{\textbf{Acc.}} &\multicolumn{1}{c}{\textbf{F1}}\\ \hline
            Static & 83.30 & 83.43 &31.45 & 28.01&96.82 & 94.67 &72.20 &71.73\\
            Dynamic  & \textbf{83.60} & \textbf{83.64}  & \textbf{32.21} & \textbf{28.91}  &\textbf{98.28} & \textbf{97.20} &\textbf{73.10} & \textbf{72.46}\\
            \bottomrule
            \end{tabular}}
        \label{tab: merging}
    \end{minipage}
\end{table}

\para{The Effect of Sparsity Control. }
We demonstrate how different levels of sparsity affect the performance of graph data augmentation. 
We control the level of sparsity using the number of edges per concept $|\mathcal{E}^{\text{conn}}|$ used for KG generation.
Table \ref{tab: num-concepts} presents the results of this study.
Given a fixed number of concepts, the performance improves when $|\mathcal{E}^{\text{conn}}|$ increases, demonstrating the effectiveness of graph merging.
However, when $|\mathcal{E}^{\text{conn}}|$ is too large compared to the original graph size, the augmented graph would be biased from too many noisy connections, and hence the observed performance deteriorates.

\para{The Influence of Different Granularity and Instruction Fine-tuning.}
We evaluate the influence of different granularity and instruction fine-tuning (IFT) on augmentation performance.
From Table \ref{tab: granularity+IFT}, it is observed that the performance is improved when an appropriate $s$ is chosen, while adopting a multi-granularity $(s_0+s_1)$ could potentially lead to over-sparsification.
With KG concepts pruned by IFT, the performance is consistently improved on different granularity levels.

\begin{figure}[H]
\centering
\begin{minipage}{0.45\textwidth}
    \captionof{table}{Performance of node classification (using GAT) with different numbers of edges per concept generated by the KG.}
    \centering
    \scalebox{0.68}{
        \begin{tabular}{l|cc|cc|cc|cc}
        \toprule
        \multicolumn{1}{l|}{} & \multicolumn{2}{c|}{\textbf{Cora}} & \multicolumn{2}{c|}{\textbf{Actor}}  & \multicolumn{2}{c|}{\textbf{PPI}} & \multicolumn{2}{c}{\textbf{Citeseer}}\\
        \multicolumn{1}{l|}{$|\mathcal{E}^{\text{conn}}|$} &
        \multicolumn{1}{c}{\textbf{Acc.}} &
        \multicolumn{1}{c|}{\textbf{F1}} &
        \multicolumn{1}{c}{\textbf{Acc.}} &
        \multicolumn{1}{c|}{\textbf{F1}} &
        \multicolumn{1}{c}{\textbf{Acc.}} &
        \multicolumn{1}{c|}{\textbf{F1}} & 
        \multicolumn{1}{c}{\textbf{Acc.}} &
        \multicolumn{1}{c}{\textbf{F1}} 
        \\ \hline
        0 &  81.3 & 80.7 & 30.3 & 28.0 &91.6 &97.1 & 72.1 &69.6 \\
        3 & 83.6 & 83.6 & 32.2 & 28.9 &96.4 &97.2 &73.1 &72.5\\
        30 & 79.3 &79.4 & 31.0 & 28.8 &97.5  &97.2 &68.5 & 68.3\\
        100 & 75.4 & 75.5 & 30.9 & 28.4 &98.3 &97.2 & 66.2 & 66.2\\
        \bottomrule
        \end{tabular}}
    \label{tab: num-concepts}
\end{minipage}
\hfill
\begin{minipage}{0.45\textwidth}
    \captionof{table}{Performance of our framework on Cora node classification with different granularity levels $s$, and with or without IFT, respectively. We denote $s_1$ as the class type level, $s_0$ as the dataset level, and $s_1+s_0$ as a multi-granularity scheme merging these two levels.}
    \centering
    \scalebox{0.75}{
        \begin{tabular}{l|cc|cc|cc}
        \toprule
        \multicolumn{1}{l|}{} & \multicolumn{2}{c|}{\textbf{$s = s_1$}} & \multicolumn{2}{c|}{\textbf{$s = s_0$}} & \multicolumn{2}{c}{\textbf{$s = s_0+s_1$}}\\
        \multicolumn{1}{l|}{IFT} &
        \multicolumn{1}{c}{\textbf{Acc.}} &
        \multicolumn{1}{c|}{\textbf{F1}} &
        \multicolumn{1}{c}{\textbf{Acc.}}&
        \multicolumn{1}{c|}{\textbf{F1}} &
        \multicolumn{1}{c}{\textbf{Acc.}} &
        \multicolumn{1}{c}{\textbf{F1}}
        \\ \hline
        w/o IFT & 81.40 & 81.53 & 82.17&82.05 & 81.00  & 81.07 \\
        w/ IFT & 83.20 & 83.26 &\textbf{83.60} &\textbf{83.64} & 83.15 & 83.25 \\
        \bottomrule
        \end{tabular}}
    \label{tab: granularity+IFT}
\end{minipage}
\end{figure}

\section{Conclusion}
We propose a novel framework for graph data augmentation, namely DemoGraph, which leverages the open-world knowledge in LLMs to perform context-driven graph data augmentation.
Our method directly operates on knowledge graphs constructed from LLM outputs and does not require access to model weights and features, which enables democratization to most of the closed-access LLMs.
To tackle the sparsity induced by generated knowledge graphs, we design a granularity-aware prompting strategy to control the sparsity while maximizing the utility of domain knowledge.
Experiments on generic graph datasets and a medical records dataset with an array of GNN architectures validate that our method can better augment the graph data than existing methods.
Ablation analysis on key components and hyperparameters of our method validates the significance of our method and robustness to variations.
Our method also has a wide range of potential application fields beyond medical record analysis such as molecular chemistry, recommendation, computational biology, social networks, and citation networks etc.

\para{Code and Data Availability.}

The codes for reproducing this work are available at \url{https://github.com/ys-feng/DemoGraph}.

\para{Acknowledgement.}
This work was supported in part by the Research Grants Council of Hong Kong (27206123 and T45-401/22-N), in part by the Hong Kong Innovation and Technology Fund (ITS/273/22 and ITS/274/22), in part by the National Natural Science Foundation of China (No. 62201483), and in part by Guangdong Natural Science Fund (No. 2024A1515011875).

\para{Declaration of Generative AI and AI-assisted Technologies in the Writing Process.}

During the preparation of this work, the author(s) used ChatGPT in order to polish the language. After using this tool/service, the authors reviewed and edited the content as needed and take full responsibility for the content of the publication.

\clearpage

\bibliographystyle{elsarticle-num} 
\bibliography{references}

\newpage
\appendix

\section{Broader Impact and Limitations}

\textbf{Border Impacts. }
The DemoGraph framework offers significant extensibility across diverse applications due to its versatile core methodologies, notably in Computational Biology, Computer Vision, and sequence data. 

In computational biology, it enhances drug discovery and protein structure prediction by generating biologically plausible augmentations for protein graphs, leveraging domain knowledge about amino acid sequences and protein interactions \citep{jiang2021could,toti2022fgdb,jumper2021highly}. This addresses the challenge of relying on vast, high-quality datasets \citep{schauperl2022ai}. 

For histopathology analysis, DemoGraph collaborates well with typical multiple-instance learning methods and can integrate the biological and clinical information described in the clinical reports with the LLM components \cite{li2021dsmil, ilse2018abmil, shao2021transmil}.

In recommendation systems, DemoGraph can facilitate graph-based methods as well as collaborative filtering (CF) methods which mitigate the bias inherent from the noisy user-iterm interaction. The use of LLM can effectively mine the contextual information from item descriptions to provide more accurate recommendations to users \cite{kojima2020kgcn, zhao2023CausRec, ji2015transD, shi2018HeRec, yang2023molerec, nguyen2018d2drec}.

\textbf{Limitations. }
Since our method operates on latent knowledge graphs, it is difficult to generate KGs and perform instruction fine-tuning online in general scenarios due to the limitations of computational resources.
However, under sufficient computational power, the LLM can be updated simultaneously during GNN training via instruction fine-tuning (e.g., after every backpropagation step) such that the generated KGs can be closer to the domain context, which is a promising extension in future works.

\section{Additional Dataset Information}

We present additional information on the datasets used in the experiments. 
Tables \ref{tab: mimic3_summary} and  \ref{tab: datasets general graphs}  present the summary information on generic graph datasets and the MIMIC-III dataset.

\begin{itemize}
\item{Cora Dataset:} The Cora dataset includes 2,708 scientific publications across seven classes, forming a citation network with 5,429 links. Each publication is represented by a binary word vector indicating the presence or absence of 1,433 unique words.

\item{Protein-Protein Interaction (PPI) Dataset:} The PPI dataset consists of graphs representing interactions between proteins in various human tissues. Nodes reflect gene expressions, and edges denote protein interactions.

\end{itemize}

\begin{table*}[ht]
\caption{Summary of the generic graph benchmark datasets.}
\centering
\begin{tabular}{lccccc}
\toprule
 & PPI & Actor & Cora & Citeseer \\
\hline
Task  & Inductive & Inductive & Transductive & Transductive \\
Nodes & 56,944 & 7,600 & 2,708 & 3,327 \\
Edges & 818,716 & 33,391 & 5,429 & 4,732 \\
Features & 50 & 932 & 1,433 & 3,703 \\
Classes & 121 & 5 & 7 & 6 \\
Training Nodes & 44,906 & 3,648 & 140 & 120 \\
Validation Nodes & 6,514 & 2,432 & 500 & 500 \\
Testing Nodes & 5,524 & 1,520 & 1,000 & 1,000 \\
\bottomrule
\end{tabular}
\label{tab: datasets general graphs}
\end{table*}

\begin{table}[h]
\centering
\caption{Summary of the OGBN datasets}
\scalebox{0.9}{
\begin{tabular}{lcccl}
\toprule
\textbf{Datasets} & \textbf{Scale} & \textbf{\# Node} & \textbf{\# Edges} & \textbf{\# Class} \\ 
\hline
OGBN-products & Large & 2,449,029 & 61,859,140 & 47 \\
OGBN-arxiv & Large & 169,343 & 1,166,243 & 40 \\
\bottomrule
\end{tabular}
}
\label{tab:OGBN_summary}
\end{table}

\begin{table*}[h]
    \caption{Summary of the MIMIC-III dataset.}
    \centering
        \begin{tabular}{lrc|lr p{1.2cm}}
        \toprule
        \multicolumn{1}{l}{\textbf{Node Type}} &
        \multicolumn{1}{c}{\textbf{Count}} &
        \multicolumn{1}{c}{\textbf{Avg. \# Visits Per Entity}} &
        \multicolumn{1}{l}{\textbf{Task}} &
                \multicolumn{1}{c}{\textbf{\# Obs.}} 
        \\ \hline
        Patients  &   46,520 & --- & Mortality & 9,718 \\
        Visits &  58,976 & --- & Readmission & 9,718 \\
        Diagnoses &  6,984 & 11.04 & LOS & 44,407\\ 
        Prescriptions &  4,204 & 70.40 & Drug Recomm. & 14,142\\
        Procedures & 2,032 & 1.55 & & \\ 
        \bottomrule
        \end{tabular}
    \label{tab: mimic3_summary}
\end{table*}

\begin{table*}[h]
\centering
\caption{Analysis of time complexity of training time on the \texttt{ogbn-arxiv} dataset.}
\label{tab:time_complexity}
\scalebox{0.75}{
\begin{tabular}{l|c|c|c|c}
\toprule
\textbf{Method} & \textbf{CGA (Ours) on GAT} & \textbf{TAPE} & \textbf{GraphGPT-stage-2} & \textbf{GraphGPT-stage-1} \\
\midrule
\textbf{Training Time (min)} & 89 & 192 & 224 & 1325 \\
\bottomrule
\end{tabular}
}
\end{table*}

\section{Evaluation Metrics}
We provide detailed definitions of the evaluation metrics.
For multi-class and multi-label classification tasks, the weighted averaging method is adopted for some metrics.
\begin{itemize}
\item Classification metrics:
\begin{itemize}
    \item Accuracy: the fraction of correct predictions to the total number of ground truth labels.
    \item F-1 score: The F-1 score for each class is defined as
    \begin{align*}
        \text{F-1 score} = 2 \cdot \dfrac{\text{precision} \cdot \text{recall}}{\text{precision} + \text{recall}}
    \end{align*}
    where `recall' is the fraction of correct predictions to the total number of ground truths in each class and precision is the fraction of correct predictions to the total number of predictions in each class. 
    \item AUC: the area under the receiver operating curve (ROC) which is the plot of the true positive rate (TPR/Recall) against the false positive rate (FPR).
    \item AUPR: the area under the precision-recall curve.
    \item Jaccard index: measures the similarity between the true binary labels and the predicted binary labels by the ratio of the size of the intersection of the true positive labels and the predicted positive labels to the size of the union of the true positive (TP) labels and the predicted positive labels including TP, false postive (FP) and false negative (FN),
    \begin{align*}
        \text{Jaccard} = \dfrac{\text{TP}}{\text{TP} + \text{FP} + \text{FN}}.
    \end{align*}
\end{itemize}
\end{itemize}

\section{Additional Information of Related Methods}
We provide supplementary information on the baseline methods and related works employed in our study. 
All baseline models were trained for 50 epochs with the option of early stopping.
We choose the model at the epoch where it reaches the best performance in terms of AUROC. 
The following is a summary of the EHR analysis included and compared:

\begin{itemize}
    \item Dipole \citep{ma2017dipole}: adopts bidirectional recurrent neural networks and attention mechanism to learn medical code representation and provide predictions.
    \item KAME \citep{ma2018kame}: a generalized version of GRAM \citep{choi2017gram} adding attention mechanism to graph representation learning to provide interpretative diagnoses.
    \item SparcNet \cite{jing2023sparcnet}: an algorithm that can classify seizures and other seizure-like events with expert-level reliability by analyzing electroencephalograms (EEGs).
    \item GRU \citep{medsker2001RNN}: a vanilla Gated recurrent unit model for visits sequence modelling.
    \item Transformer \citep{vaswani2017transformer}: it leverages the idea of self-attention, which allows the model to selectively focus on different parts of the input sequence when generating an output. 
    \item GRAM \citep{choi2017gram}: the first work models EHR with a knowledge graph and uses recurrent neural networks to learn the medical code representations and predict the future visit information. 
    \item StageNet \citep{gao2020stagenet}: using a stage-aware LSTM to conduct clinical predictive tasks while learning patient disease progression stage change in an unsupervised manner.
    \item Concare \citep{ma2020concare}: it considers personal characteristics during clinical visits and uses cross-head decorrelation to capture inter-dependencies among dynamic features and static baseline information for predicting patients' clinical outcomes given EHRs.
    \item Adacare \citep{ma2020adacare}: it can capture the long and short-term variations of biomarkers as clinical features, model the correlation between clinical features to enhance the ones which strongly indicate health status, and provide qualitative interpretability while maintaining a state-of-the-art performance in terms of prediction accuracy.
    \item Dr. Agent \citep{gao2020dragent}: mimics clinical second opinions using two reinforcement learning agents and learn patient embeddings with the agents.
    \item GRASP \citep{zhang2021grasp}: applies GNN to cluster patients using their latent features and identifies similar patients based on latent clusters. 
    \item GraphCare \citep{jiang2023graphcare}: integrates external open-world knowledge graphs (KGs) into the patient-specific KGs with large language models.
\end{itemize}

For drug recommendation, we also included the following additional competitors.

\begin{itemize}
    \item MICRON \citep{yang2021micron}: a sophisticated personalized drug recommendation system that incorporates patient-specific genetic and molecular information. It utilizes multi-omics data to identify optimal drug combinations, especially for complex diseases, thereby increasing treatment specificity based on patients' molecular characteristics.
    \item Safedrug \citep{yang2021safedrug}: a drug-drug-interaction-controllable (DDI-controllable) drug recommendation model that leverages drugs' molecule structures and models DDIs explicitly. It uses a global message passing neural network (MPNN) module and a local bipartite learning module to fully encode the connectivity and functionality of drug molecules.
    \item MoleRec \citep{yang2023molerec}: a novel molecular substructure-aware encoding method that employs a hierarchical architecture to model inter-substructure interactions and the impact of individual substructures on a patient’s health condition.
\end{itemize}

\section{Implementation Details}
\para{Temperature annealing. } We are aware of the vanishing classification loss in practice. Therefore, we alleviate this issue by annealing the temperature over the training epochs with the schedule $\tau = \max (0.5, \exp(rp))$, where $p$ is the training epoch and $r = 0.01$.

\para{Downsampling for mortality task. }
We are aware that the samples in the mortality prediction task are heavily imbalanced (i.e., most of the samples are not dead).
We therefore perform downsampling during training to balance the samples.

\para{Configurations. }
The proposed framework is implemented in Python with the \textit{Pytorch} library on a server equipped with four NVIDIA GeForce RTX 3090 GPUs. 
We use the \textit{dgl} library to perform graph-related operations, and \textit{pyhealth} \citep{zhao2021pyhealth} to benchmark SOTA methods and perform EHR-related operations.
We integrate the gpt-4-0125-preview API \citep{openai2023gpt4}, serving as a frozen large language model.
The dropout ratio of each dropout layer is set as 0.1. 
All models are trained with 1000 epochs with early stopping. 
We use the Adam optimizer to optimize the model with a learning rate of $5 \times 10^{-5}$ and a weight decay of $1 \times 10^{-5}$. 

\section{Compute Amount Analysis}

\para{Time Complexity Analysis. }
Since we generate the KGs offline using the OpenAI API of gpt-4-0125-preview (our method works under a black-box setting), this process only needs to be performed once for each dataset. The additional complexity arises from the dynamic merging process, which needs to be repeated at each optimization step. However, the time complexity of this step is trivial compared to the forward passing of GNNs. Therefore, it only increases the overall time complexity on a minor level.

Table \ref{tab:time_complexity} above shows the quantitative analysis of the training time complexity on the \texttt{ogbn-arxiv} dataset.

\para{Efficiency through Single Query and Reuse. }
our prompting paradigm avoids manual prompt customization for adaptations to different datasets, thereby reducing human labor costs. Our method necessitates only a single query to the LLM, with KGs and significant concept nodes stored for subsequent reuse. Our query process can be efficiently completed in 37.6 seconds in average for the large-scale ogbn-arxiv dataset. This approach not only enhances efficiency but also reduces the number of API calls, thereby saving the cost of commercial LLMs. Additionally, we have provided the responses from the LLMs gained in our experiments for the public use.

\section{Additional Details on Healthcare Tasks}

We include detailed descriptions of the healthcare tasks performed on the EHR datasets.

\para{Mortality Prediction}
Mortality prediction aims to predict the mortality label of the subsequent visit for each sample. Formally, the function \( f: (x_1, x_2, \ldots, x_{t-1}) \rightarrow y[x_t] \) is defined, where \( y[x_t] \in \{0, 1\} \) is a binary label indicating the patient's survival status recorded in visit \( x_t \).

\para{Readmission Prediction}
The task of readmission prediction focuses on whether a patient will be readmitted within \( \delta \) days. Formally, \( f: (x_1, x_2, \ldots, x_{t-1}) \rightarrow y[\delta(x_t) - \delta(x_{t-1})] \), where \( \delta(x_t) \) represents the encounter time of visit \( x_t \), so that \( y[\delta(x_t) - \delta(x_{t-1})] \) is 1 if \( \delta(x_t) - \delta(x_{t-1}) \leq \delta \) and 0 otherwise. For our EHR study, we set \( \delta = 15 \) days.

\para{Length-Of-Stay Prediction}
Length-Of-Stay (LOS) prediction predicts the length of ICU stay for each visit. Formally, \( f: (x_1, x_2, \ldots, x_t) \rightarrow y[x_t] \), where \( y[x_t] \in \mathbb{R}^{1 \times C} \) is a one-hot vector indicating its class among \( C \) classes.

\para{Drug Recommendation}
This task predicts medication labels for each visit. Formally, \( f: (x_1, x_2, \ldots, x_t) \rightarrow y[x_t] \), where \( y[x_t] \in \mathbb{R}^{1 \times |d|} \) is a multi-hot vector where \( |d| \) denotes the number of all drug types.

\section{Additional Experiment Results}

We present additional experiment results with more metrics for comparisons.
Table \ref{tab: results_MIMIC3_drug_LOS_full}  and \ref{tab: results_MIMIC3_Drug_full} present the performance on length of stay prediction and drug recommendations with more evaluation metrics. 

\begin{table}[]
    \caption{Performance (in \%) of our method on the drug recommendation task on the MIMIC-III dataset. Standard deviations are shown in brackets. }
    \centering
    \scalebox{0.8}{
        \begin{tabular}{lcccc}
 & \multicolumn{4}{c}{\textbf{Drug Recommendation}}\\
        \toprule
        \multicolumn{1}{c}{\textbf{Model}} &
        \multicolumn{1}{c}{\textbf{AUPRC}} &
        \multicolumn{1}{c}{\textbf{AUROC}} &
        \multicolumn{1}{c}{\textbf{F1-score}} &
        \multicolumn{1}{c}{\textbf{Jaccard}} \\ \hline
        GRU & 77.0(0.1) & 94.4(0.0) & 62.3(0.3) & 47.8(0.3)  \\
        Transformer & 76.1(0.1) & 94.2(0.0) & 62.1(0.4) & 47.1(0.4)  \\
        RETAIN & 77.1(0.1) & 94.4(0.0) & 63.7(0.2) & 48.8(0.2) \\
        GRAM & 76.7(0.1) & 94.2(0.1) & 62.9(0.3) & 47.9(0.3)  \\
        Deepr & 74.3(0.1) & 93.7(0.0) & 60.3(0.4) & 44.7(0.3) \\
        StageNet & 74.4(0.1) & 93.0(0.1) & 61.4(0.3) & 45.8(0.4) \\
        SafeDrug & 68.1(0.3) & 91.0(0.1) & 46.7(0.4) & 31.7(0.3)  \\
        MICRON & 77.4(0.0) & 94.6(0.1) & 63.2(0.4) & 48.3(0.4) \\
        GAMENet & 76.4(0.0) & 94.2(0.1) & 62.1(0.1) & 47.2(0.4) \\
        MoleRec & 69.8(0.1) & 92.0(0.1) & 58.1(0.1) & 43.1(0.3) \\
        GraphCare& 78.5(0.2) & 94.8(0.1) & 64.4(0.3) & 49.2(0.4) \\
        Ours& \textbf{81.8(0.1) }&\textbf{ 97.1 (0.2)}&\textbf{ 66.1 (0.2)} & \textbf{49.4 (0.8)} \\ \hline
        \bottomrule
        \end{tabular}}
    \label{tab: results_MIMIC3_drug_LOS_full}
\end{table}

\begin{table}[]
    \caption{Performance (in \%)  of our method on prediction of the length of stay on the MIMIC-III datasets. Standard Deviations are shown in brackets.}
    \centering
    \resizebox{0.9\linewidth}{!}{%
        \begin{tabular}{lccc p{1.3cm}}
        \toprule
        \multicolumn{1}{c}{}& \multicolumn{3}{c}{\textbf{Prediction of Length of Stay}} \\
        \multicolumn{1}{c}{}&
        \multicolumn{3}{c}{\textbf{MIMIC-III}}  \\
        \multicolumn{1}{l}{\textbf{Model}} &
        \multicolumn{1}{c}{\textbf{Accuracy}} &
        \multicolumn{1}{c}{\textbf{AUROC}}& 
        \multicolumn{1}{c}{\textbf{F1}} \\ \hline
        GRU&  42.14 (0.6) & 80.23 (0.2)  &  27.36 (0.7)  \\
        Transformer & 41.68 (0.7) & 79.30 (0.8) & 27.52 (0.8)  \\
        Deepr  & 39.31 (1.2) & 78.02 (0.4) & 25.09 (1.3)  \\
        GRAM & 40.00  (0.0) & 78.00  (0.0) & 34.00 (0.0)  \\
        Concare  & 42.04 (0.6) & 80.27 (0.3) & 25.44 (1.3) \\
        Dr. Agent & 41.40 (0.5) & 79.45 (0.6) & 27.55 (0.3) \\ 
        AdaCare  & 40.7 (0.8) & 78.73 (0.4) & 26.26 (0.8) \\ 
        StageNet & 40.18 (0.7) &  77.94 (0.2)  & 26.63 (1.2)  \\
        GRASP & 40.66 (0.3) &  78.97 (0.4) & 22.80 (0.8)  \\
        GraphCare  & 43.20 (0.4) & 81.40 (0.3) & 37.50 (0.2) \\
        \multicolumn{1}{l}{\textbf{DemoGraph}}&46.28 (1.0) &85.68 (0.1) & 38.67 (0.6)\\ \hline
        \bottomrule
        \end{tabular}
    }    
    \label{tab: results_MIMIC3_Drug_full}
\end{table}

\section{Additional Details on GNN Architectures}

\para{Graph Neural Network (GNN).}
A GNN, denoted as $\mathcal{M}$, operates on $\mathcal{G}$ and takes its feature space $\mathcal{X}$ to perform prediction by message passing. 
The message-passing mechanism of GNNs can be presented as
\begin{align*}
h_v^{(l+1)} = \text{UPDATE}^{(l)} \left( h_v^{(l)}, \text{AGG}^{(l)} \big\{ h_u^{(l)}: u \in \mathcal{N}(v) \big\}  \right),
\end{align*}
where \( h_v^{(l)} \) is the feature vector of node \( v \) at the \( l \)-th layer, \( \mathcal{N}(v) \) denotes the set of neighbors of \( v \), and \text{UPDATE(•)} and \text{AGG(•)} are functions for the update and aggregation steps respectively.

\begin{itemize}
    \item Graph Convolutional Network (GCN): operates on graphs using a spectral approach for convolution, aggregating neighbor node information to update node features.
    \item Graph Attention Network (GAT): utilizes attention mechanisms to weigh neighbor contributions, employing a self-attention mechanism for attention weight calculation and neighbor feature aggregation.
    \item Graph Isomorphism Network (GIN): aggregates neighbor features using a learnable function, maintaining invariance to neighbor ordering in both directed and undirected graphs.
    \item GraphSAGE \citep{hamilton2017graphSAGE}: a pioneer sampling and aggregating algorithm for inductive graph representation learning.
\end{itemize}

\section{Additional Ablation Studies} 

\para{The Influence of Number of GNN Layers. }
We evaluate the performance of our method with different numbers of GNN layers, as summarized in
Table \ref{tab: no-layers}.
We observe that in general a better performance is obtained when the number of layers is small.
The performance slightly deteriorates as the number of layers increases more than two layers, indicating the potential over-smoothing problem.
Other experiments on relatively fine-grained hyperparameters, such as the dropout rate, number of hidden dimensions, and number of attention heads for GAT, are presented in the appendix. 
\begin{table}[h]
    \caption{
    Performance in terms of accuracy (\%) of our framework on node classification with different numbers of layers $L$, using  GCN and GAT. 
    }
    \centering
    \scalebox{0.8}{
        \begin{tabular}{l|ccc|cccp{1.2cm}}
        \toprule
        \multicolumn{1}{l|}{} & \multicolumn{3}{c|}{\textbf{GCN}}  & \multicolumn{3}{c}{\textbf{GAT}}\\
        \multicolumn{1}{l|}{$L$} & \multicolumn{1}{c}{\textbf{Cora}} & \multicolumn{1}{c}{\textbf{Actor}} &  \multicolumn{1}{c|}{\textbf{Citeseer}}   & \multicolumn{1}{c}{\textbf{Cora}} & \multicolumn{1}{c}{\textbf{Actor}} &  \multicolumn{1}{c}{\textbf{Citeseer}}
        \\ \hline
        1 & 81.40 & 31.91 & 71.77  & 83.30  & 32.21 & 72.70\\
        2 & 81.50 & 32.41 & 73.10 & 83.60& 29.21 & 73.10\\
        3 & 82.90 & 31.45 & 70.45 & 82.10 & 28.49 & 72.10\\
        4 & 80.50 & 30.54 & 70.04 & 81.70 & 28.20 & 71.70\\
        \bottomrule
        \end{tabular}}
    \label{tab: no-layers}
\end{table}

\begin{figure}
    \centering
    \includegraphics[width=1\linewidth]{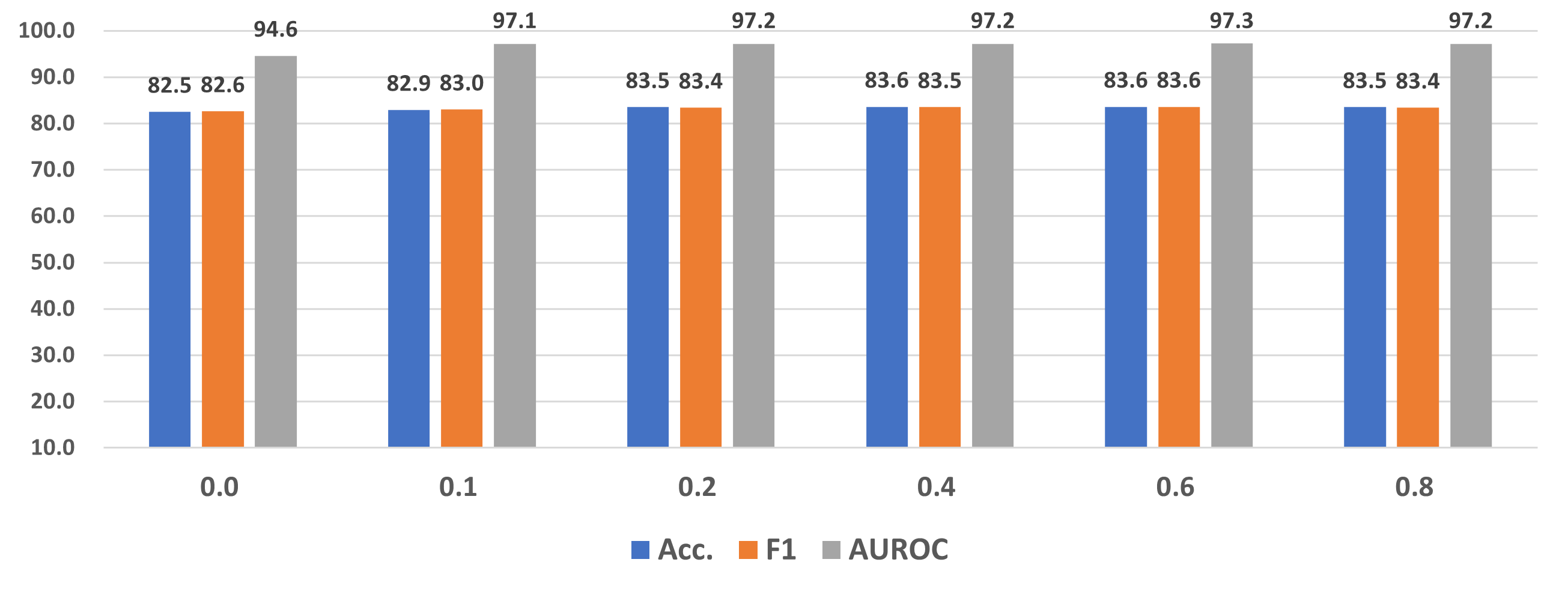}
    \caption{Performance of our method on Cora node classifications with respect to different dropout ratios, with GAT as the GNN architecture.}
    \label{fig: abl-drops}
\end{figure}

\para{Dropout Ratios.}
Since graph learning is difficult to optimize and easy to lead to overfitting, we adopt dropout as the default regularizer for all benchmark methods.
We further study the effects of different dropout rates, Figure \ref{fig: abl-drops} presents the results.
We observe that our method is in general robust to changes in dropout rates while being optimized when the dropout rate is 0.6.
However, a large dropout rate would lead to over-sparsification of neural network weights and important features being dropped, hindering the predictive performance.

\begin{figure}[h]
    \centering
    \includegraphics[width=1\linewidth]{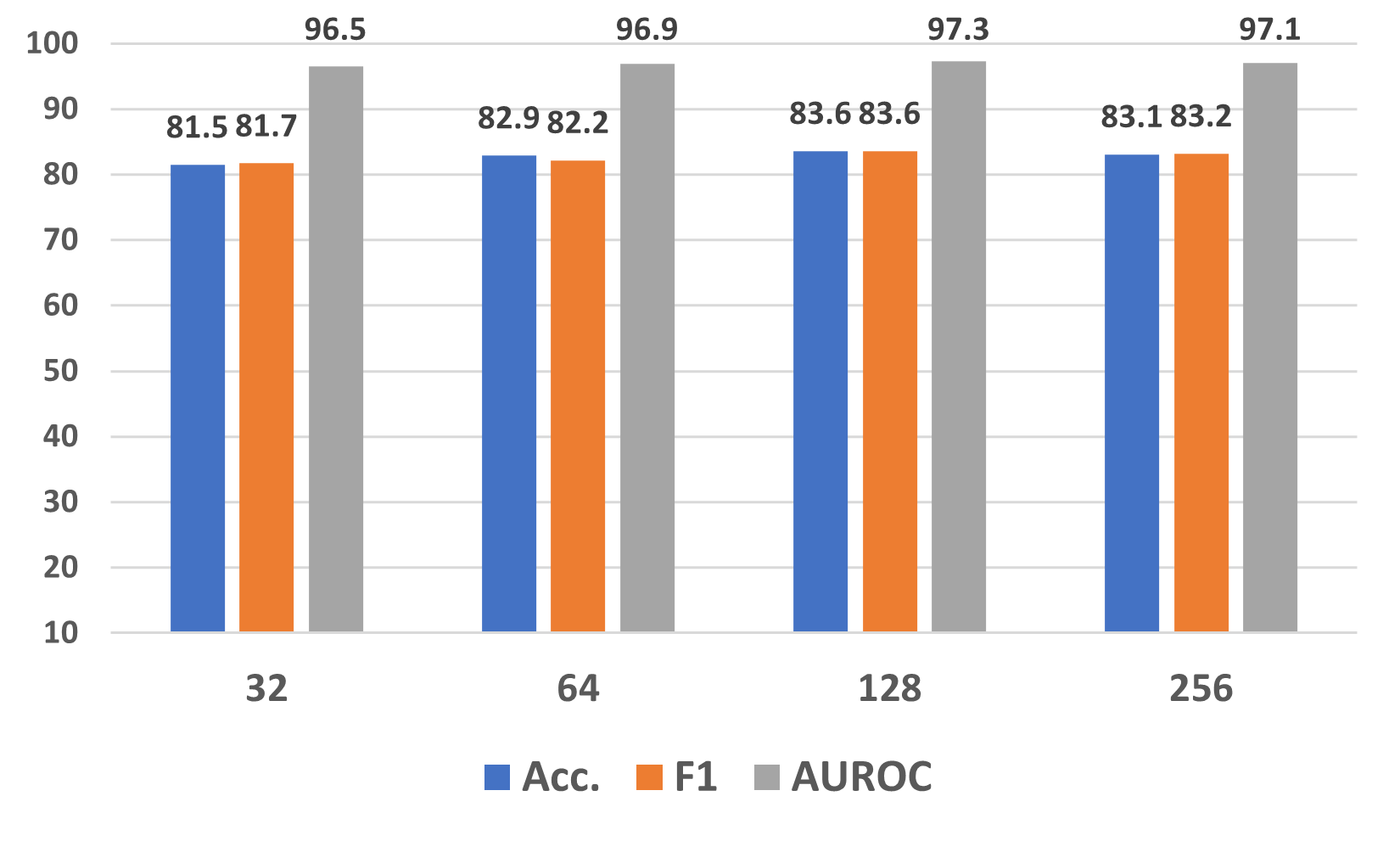}
    \caption{Performance of our method on Cora node classifications with respect to different numbers of hidden dimensions, with GAT as the GNN architecture.}
    \label{fig: abl-dimensions}
\end{figure}

\para{ Number of hidden dimensions. } We benchmark our method with respect to different hidden dimensions.
Figure \ref{fig: abl-dimensions} presents the results of this study.
\begin{figure}[h]
    \centering
    \includegraphics[width=1\linewidth]{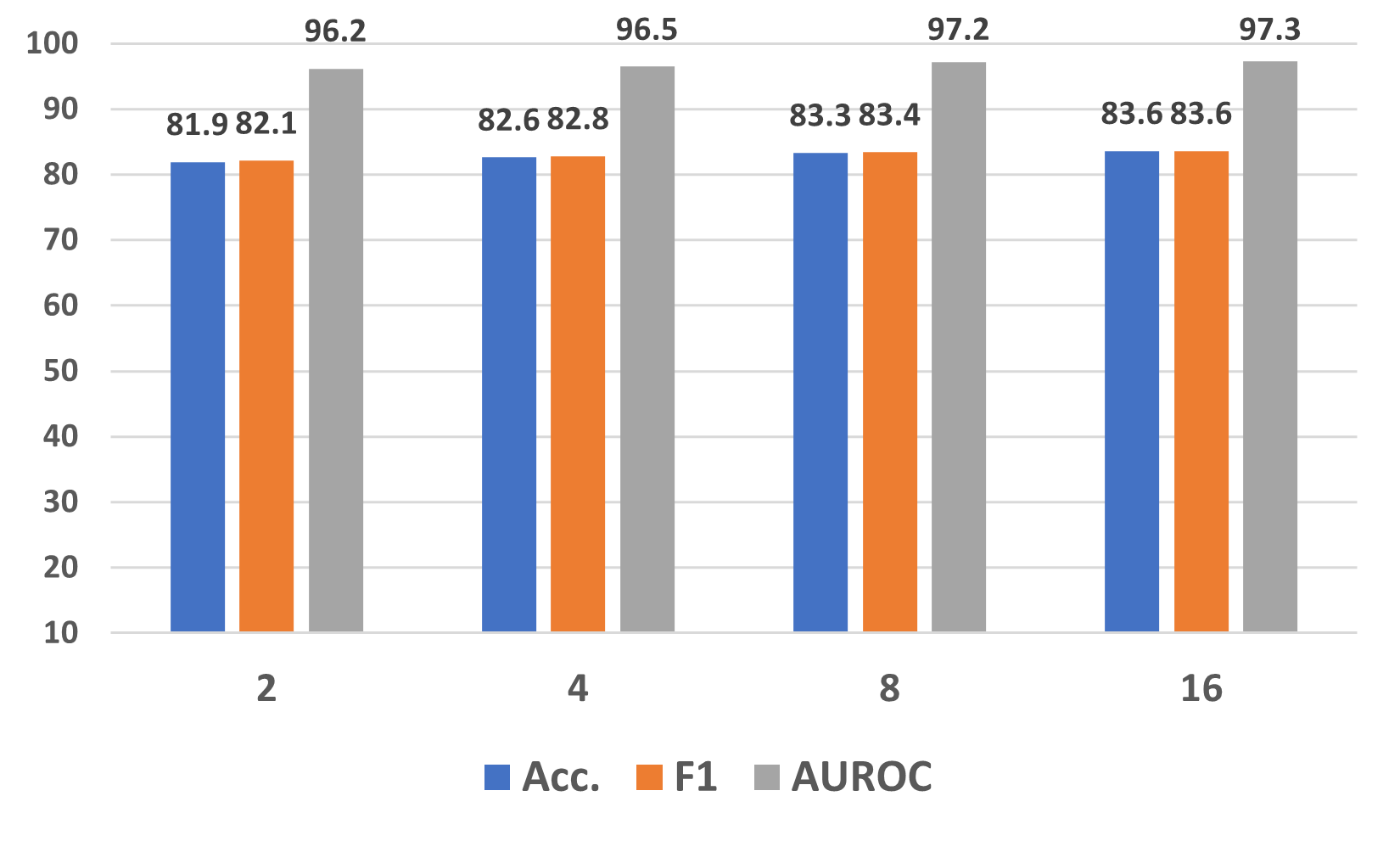}
    \caption{Performance of our method on Cora node classifications with respect to different heads of GAT. }
    \label{fig: abl-heads}
\end{figure}
We observe our method is overall robust to different numbers of hidden dimensions.
In general, a larger number of hidden dimensions leads to better classification performance. 

\para{ Number of attention heads}. We benchmark the performance of our method with respect to different attention heads, as summarized in Figure \ref{fig: abl-heads}.
We observe that the performance is overall improving with the number of heads increases, while a larger number of heads (e.g., 32) would lead to a heavier memory burden under the current hardware settings. 

\section{Details of Prompting Designs}

We present additional examples of our prompting designs, including the ones used in KG generations for generic graph datasets (Cora, Citeseer, PPI, Actor).

\para{Prompt Template for Cora:}
Specifically, for the Cora dataset of scientific publications, we follow the structure mentioned before and thus design an effective prompt as follows.
\begin{lstlisting}[]
Given a prompt (a node from Cora dataset with Neural_Networks as label), generate an extensive array of associated connections based on your domain knowledge, which shuold be helpful for the "node classification task." [Replaced with the description of the specific downstream task(s).]
Note that the updates should based on the provided node label and important indices and backed up by your knowledge, being reasonable and useful. It should not be unnecessary or nonsense.
"The Cora dataset consists of 2708 scientific publications classified into one of seven classes.The citation network consists of 5429 links. Each publication in the dataset is described by a 0/1-valued word vector indicating the absence/presence of the corresponding word from the dictionary. The dictionary consists of 1433 unique words." [Replaced with the description of the specific dataset.]
Format each association as [ENTITY 1, RELATIONSHIP, ENTITY 2], ensuring the sequence reflects the direction of the relationship. Both ENTITY 1 and ENTITY 2 are to be nouns. Elements within [ENTITY 1, RELATIONSHIP, ENTITY 2] must be definitive and succinct.
{example}
prompt: f"Node {term} in Cora with label {mode} and important indices {non_zero_indices}"
updates:
\end{lstlisting}

\para{Prompt Example for Citeseer (a scientific publications citation dataset) on the Granularity Level $s_0$ with IFT:}
Specifically, for the Citeseer dataset of citation networks of scientific publications, we provide a concrete example of prompts following the prompting strategy discussed before.

\begin{lstlisting}[]
Given the The Citeseer dataset,which consists of 3312 scientific publications classified into one of six classes. The citation network consists of 4732 links. Each publication in the dataset is described by a 0/1-valued word vector indicating the absence/presence of the corresponding word from the dictionary. The dictionary consists of 3703 unique words.
Please generate 100 important concepts that related to the whole dataset and all subtypes(Agents, Artificial Intelligence, Database, Human Computer Interaction, Machine Learning and Information Retrieval.), which are crucial for downstream task like node classification.
Each concept should be a single term or a short phrase that encapsulates an important idea, technique, or subject within these domains.
Make sure the concepts are relevant to the whole dataset and could  improve the downstream tasks' performance.
\end{lstlisting}

The following prompt are for the instruction fine-tuning process for concept pruning:

\begin{lstlisting}[]
Among the prior 100 concepts, select 30 most important concepts from the list. The importance of a concept is based on your knowledge and inference on how it will help improve the node classification task on Citeseer dataset. If you think a concept is important, please keep it. Otherwise, please remove it.
\end{lstlisting}

The following prompt is for the KG generation given the pruned concepts output from the LLM using the previous prompts.

\begin{lstlisting}[]
You are now a professional scientific publication researcher. Given the list of 30 important concepts augmented with the Citeseer datset, extrapolate 100 relationships of it and provide a list of triples.
{context_descriptions}
The relationships should be helpful for downstream node classification task.
Each update should be exactly in format of [ENTITY 1, RELATIONSHIP, ENTITY 2]. The relationship is directed, so the order matters.
Both ENTITY 1 and ENTITY 2 should be selected from the list of 30 important concepts.
The selection of entities and the construction of relationship should be supported by your professional domain knowledge, and do make sense. It should not be nonsense or randomly constructed.
Consideting the complexity of the task, you can iterate each one of the concept in the list and consider them as 'ENTITY1', and provide the name of 'ENTITY2', which are also from the list, to fulfill the requirement stated before.
\end{lstlisting}
where the ``context$\_$descriptions" in $\{\}$ is prompted into the LLM in the previous prompt.

\para{Prompt Example for PPI (a protein-protein interaction dataset):}
Specifically, for the PPI dataset of protein-protein interaction in the Computational Biology field, we provide a concrete example of prompts following the prompting strategy discussed before.
\begin{lstlisting}[]
Given the The PPI dataset,which is protein-protein interaction network that contains physical interactions between proteins that are experimentally documented in humans, such as metabolic enzyme-coupled interactions and signaling interactions. Nodes represent human proteins and edges represent physical interaction between proteins in a human cell. It contains 24 graphs. The average number of nodes per graph is 2372. Each node has 50 features and 121 labels. 20 graphs for training, 2 for validation and 2 for testing.
Please generate 200 important concepts that related to the whole Protein-protein Interactions dataset and all its subtypes, which are crucial for downstream task like node classification.
Each concept should be a single term or a short phrase that encapsulates an important idea, technique, or subject within these domains.
Make sure the concepts are relevant to the whole dataset and could improve the downstream tasks' performance.
\end{lstlisting}

The following prompt are for the instruction fine-tuning process for concept pruning:

\begin{lstlisting}[]
Among the prior 200 concepts, select the 100 most important concepts from the list. The importance of a concept is based on your knowledge and inference of how it will help improve the node classification task on the Actor dataset. If you think a concept is important, please keep it. Otherwise, please remove it.
\end{lstlisting}
The following prompt is for the KG generation given the pruned concepts output from the LLM using the previous prompts.
\begin{lstlisting}[]
You are now a professional computational biology researcher. Given the list of 100 important concepts augmented with the Protein-Protein Interaction(PPI) datset, extrapolate 300 relationships between the concept nodes and provide a list of triples. 
{context_descriptions}
The relationships should be helpful for downstream node classification task on PPI dataset.
Each update should be exactly in format of [ENTITY 1, RELATIONSHIP, ENTITY 2]. The relationship is directed, so the order matters.
Both ENTITY 1 and ENTITY 2 should be selected from the list of 200 important concepts in the txt file uploaded.
The selection of entities and the construction of relationship should be supported by your professional domain knowledge, and do make sense. It should not be nonsense or randomly constructed.
\end{lstlisting}
where the ``context$\_$descriptions" in $\{\}$ is prompted into the LLM in the previous prompt.

\para{Prompt Example for Actor (a dataset of film professionals relationships) with IFT:}
Specifically, for the Actor dataset of connections among film professionals, we provide a concrete example of prompts following the prompting strategy discussed before.
\begin{lstlisting}[]
Given the Actor dataset, which is crawled from Wikipedia under the category of "English-language films". In total, there are Nodes: 7600, Edges: 33391, Number of Classes: 5. The relationship types include: film-director, film-actor, film-writer, and other relationships between actors, directors, and writers. The first three types of relationships are extracted from the "infobox" on the films' Wiki pages. All the other types of people relationships are created as follows: if one person (including actors, directors, and writers) appears on another people's page, then a directed relationship is created between them.
Please generate 100 important concepts of 'film/director/actor/writer' that are related to the whole dataset and all 5 subtypes, which are crucial for downstream tasks like node classification.
Each concept should be a single term or a short phrase that encapsulates an important idea, technique, or subject within these domains.
Make sure the concepts are relevant to the whole dataset and could improve the downstream tasks' performance.
\end{lstlisting}

The following prompt are for the instruction fine-tuning process for concept pruning:

\begin{lstlisting}[]
Among the prior 100 concepts, select the 30 most important concepts from the list. The importance of a concept is based on your knowledge and inference of how it will help improve the node classification task on the Actor dataset. If you think a concept is important, please keep it. Otherwise, please remove it.
\end{lstlisting}

The following prompt is for the KG generation given the pruned concepts output from the LLM using the previous prompts.

\begin{lstlisting}[]
You are now a professional film industry researcher. Given the list of 30 important concepts augmented with the Actor dataset, extrapolate 100 relationships between the concept nodes and provide a list of triples. {context_descriptions}
The relationships should be helpful for the downstream node classification task. Each update should be exactly in the format of [ENTITY 1, RELATIONSHIP, ENTITY 2]. The relationship is directed, so the order matters.
Both ENTITY 1 and ENTITY 2 should be selected from the list of 30 important concepts in the txt file uploaded.
The selection of entities and the construction of relationships should be supported by your professional domain knowledge, and do make sense. It should not be nonsense or randomly constructed.
\end{lstlisting}

where the ``context$\_$descriptions" in $\{\}$ is prompted into the LLM in the previous prompt.

\section{Differences Between Our Work and GraphCare 
}

We highlight a closely related work to ours --- GraphCare \cite{jiang2023graphcare}. Although both works borrow the knowledge from LLMs to graph learning domain,
GraphCare can be viewed as a special case of our work where patients are modelled as a personalized graph.
GraphCare distinctively requests clinical reports and enriched contextual information, making it difficult to generalize to scenarios when such information is scarce or not available.
Our method focuses on graph data augmentation in general, where our designed prompting strategy is applicable to almost all graph representation learning scenarios.
Our method also involved sparsity control designs such as granularity-aware prompting and IFT concept pruning, which are absent in the work of GraphCare.
Empirical comparisons on the EHR dataset also validate that our proposed graph data augmentation method outperforms their design.

\section{Extended Summary of Graph Data Augmentation Works}

\para{Modality-oriented Augmentation:} 
GDA includes structure/feature/label-oriented techniques based on the different types of information modalities present in graphs. Structure-oriented GDA adopted edge perturbation \cite{velivckovic2018DGinfomax}, graph diffusion \cite{topping2021understanding, zheng2020robust, qiu2020gcc, park2021metropolis}, graph sampling \cite{hamilton2017inductive, qiu2020gcc}, node dropping and insertion \cite{gilmer2017neural}, etc. Feature-oriented techniques focus on feature corruption \cite{feng2019graph, velivckovic2018DGinfomax, yang2021graph}, masking \cite{you2020graph}, rewriting \cite{wang2020nodeaug, yang2019topology}, and mixing \cite{verma2019manifold}. Label-oriented GDA directly enriches the expensive labeled data mainly by two ways: Pseudo-labeling and Label Mixing \cite{zhang2017mixup}. While these methods may be helpful for certain tasks, their granularity levels are inflexible and semantic awareness is lost.

\para{Graph Data Augmentation for Low-resource Graph Learning:} 
Graph self-supervised learning explores generative modeling and contrastive learning. Techniques like denoising link reconstruction \cite{hu2019pre} and GPT-GNN \cite{hu2020gpt} utilize edge perturbation and feature masking for data augmentation, aiming to reconstruct augmented graph features. Graph contrastive learning, exemplified by Deep Graph Infomax \cite{velivckovic2018DGinfomax} and GCC \cite{qiu2020gcc}, employs feature shuffling and graph sampling to generate contrasting graph samples.

Semi-supervised GDA enhances DGL by using unlabeled data. Key methods include self-training \cite{yarowsky1995unsupervised}, co-training \cite{blum1998combining}, imbalanced training \cite{zhao2021graphsmote}, consistency training \cite{xie2020unsupervised}, which aligns node representations between original and augmented graphs, and graph data interpolation \cite{zhang2017mixup}, creating synthetic examples through feature and label mixing.


\para{Towards Reliable Augmentation:} 
However, most existing techniques apply augmentations uniformly without considering the underlying graph semantics. This can introduce undesirable artifacts by distorting important graph structures and losing interpretability. Recent works have begun to train graphs in latent space \cite{yue2022label} or incorporate context and enhanced knowledge when perturbing graphs~\cite{wang2019kgat, jiang2023graphcare}, but generating reliable graphs conditioned on context remains an open challenge.

\clearpage

\end{document}